\documentclass[lettersize,journal]{IEEEtran}
\usepackage[table,x11names]{xcolor}
\usepackage{amsmath,amsfonts}
\usepackage{algorithmic}
\usepackage{algorithm}
\usepackage{array}
\usepackage[caption=false,font=normalsize,labelfont=sf,textfont=sf]{subfig}
\usepackage{textcomp}
\usepackage{stfloats}
\usepackage{url}
\usepackage{verbatim}
\usepackage{graphicx}
\usepackage{cite}
\usepackage{multirow}
\usepackage{marvosym} 
\usepackage[most]{tcolorbox}
\usepackage{listings}
\usepackage{graphicx}
\usepackage{amssymb}
\usepackage{paralist}
\usepackage{array}
\usepackage{xspace}
\usepackage{booktabs}
\usepackage[colorlinks,
            linkcolor=blue,
            citecolor=blue]{hyperref}

\definecolor{lightgray}{gray}{0.95}
\definecolor{deepblue}{RGB}{70,130,180}
\definecolor{deepgray}{RGB}{119,136,153}
\lstdefinestyle{prompt}{
    basicstyle=\ttfamily\fontsize{7pt}{8pt}\selectfont,
    frame=none,
    breaklines=true,
    backgroundcolor=\color{lightgray},
    breakatwhitespace=true,
    breakindent=0pt,
    escapeinside={(*@}{@*)},
    numbers=none,
    numbersep=5pt,
    xleftmargin=5pt,
    aboveskip=2pt,
    belowskip=2pt,
}
\definecolor{iccvblue}{rgb}{0.21,0.49,0.74}
\definecolor{color1}{RGB}{192,0,0}
\definecolor{color2}{RGB}{35,87,35}
\newcommand{\Paragraph}[1]{\noindent\textbf{#1}}

\newcommand{\sys}{Edit Transfer\xspace}

\def\eg{\textit{e.g.}\xspace}

\newcommand{\secref}[1]{Section~\ref{sec:#1}}
\newcommand{\subsecref}[1]{Section~\ref{subsec:#1}}
\newcommand{\figref}[1]{Fig.~\ref{fig:#1}}
\newcommand{\tabref}[1]{Table~\ref{tab:#1}}
\newcommand{\eqnref}[1]{Eq.~\eqref{eq:#1}}

\begin{document}

\title{EditTransfer++: Toward Faithful and Efficient \\
Visual-Prompt-Guided Image Editing}

\author{Lan Chen, Qi Mao, \textit{Member, IEEE}, Yiren Song, Yuchao Gu, Siwei Ma,  \textit{Fellow, IEEE}
\thanks{Lan Chen and Qi Mao are with the School of Information and Communication Engineering and the State Key Laboratory of Media Convergence and Communication, Communication University of China, Beijing 100024, China (E-mail: chenlaneva@mails.cuc.edu.cn, qimao@cuc.edu.cn) (Corresponding author: Qi Mao).
\\
Yiren Song and Yuchao Gu are with ShowLab, National University of Singapore, 119077, Singapore (E-mail: yuchaogu@u.nus.edu, yiren@nus.edu.sg).
\\
Siwei Ma is with the State Key Laboratory of Multimedia linformation
Processing, School of Computer Science, Peking University, Beijing 100871, China (E-mail:swma@pku.edu.cn).}
}

\markboth{UNDER REVIEW}{UNDER REVIEW}

\maketitle
\begin{abstract}
Visua-prompt-guided edit transfer aims to learn image transformations directly from example pairs, offering more precise and controllable editing than purely text-driven approaches. 
However, existing diffusion transformer–based methods often fail to faithfully reproduce the demonstrated edits due to structural mismatches between the task and the backbone, including a pretrained bias toward textual conditioning and inherent stochastic instability during sampling. 
To bridge this gap, we present EditTransfer++, a framework that combines progressively structured training with an efficient conditioning scheme to improve both visual prompt faithfulness and inference efficiency.
We first mitigate textual dominance with a text-decoupled training strategy that removes text conditioning during fine-tuning, compelling the model to infer transformations solely from visual evidence while still supporting optional text guidance at inference.
On top of this visually grounded model, a best--worst contrastive refinement mechanism reshapes the denoising trajectories to suppress unfaithful generations and improve consistency across random seeds. 
To alleviate the computational bottleneck of high-resolution in-context editing, we further introduce a condition compression and reuse strategy that reduces token redundancy and enables efficient generation of images with a 1024-pixel long edge. 
Extensive experiments on existing benchmarks and the proposed EditTransfer-Bench show that EditTransfer++ achieves state-of-the-art visual prompt faithfulness with substantially faster inference than prior methods, suggesting a promising direction for scalable prompt-guided image editing and broader visual in-context learning.
\end{abstract}


\section{Introduction}
\label{sec:intro}
\begin{figure}[!t]
    \includegraphics[width=\linewidth]{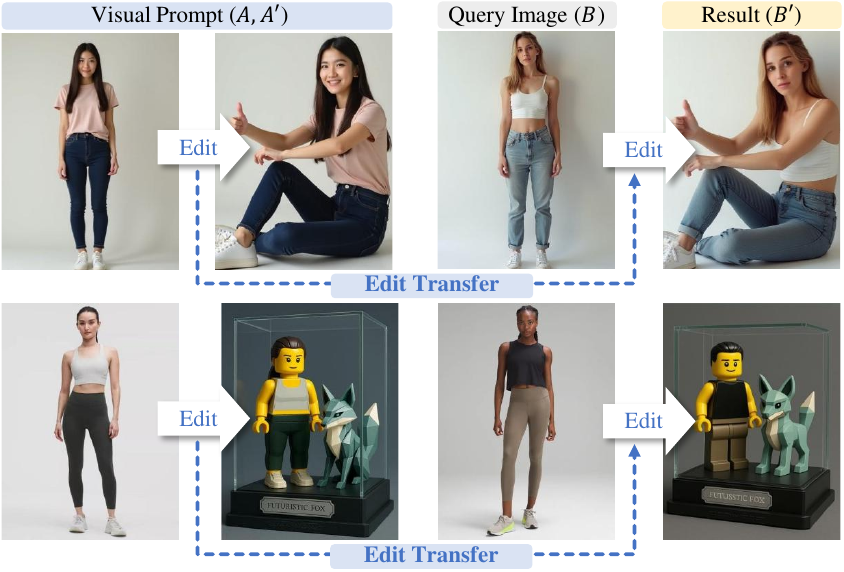}
    \vspace{-7 mm}
    \caption{\textbf{Illustration of the edit transfer task.}
A visual prompt is defined as a pair of images $(A, A')$, where $A'$ is an edited version of $A$.
Given a query image $B$, the goal of edit transfer is to apply the transformation demonstrated by $(A, A')$ to $B$, yielding an edited result $B'$.}
    \label{fig:edittransfer}
    \vspace{-4 mm}
\end{figure}

\IEEEPARstart{I}{mage} editing has rapidly advanced in recent years, with text-based image editing methods (TIE)~\cite{hertz2022prompt,cao_2023_masactrl,brooks2023instructpix2pix,zhang2025context,labs2025flux} becoming the dominant paradigm due to their flexibility in specifying user intent through natural language.
However, textual descriptions often fail to convey fine-grained transformations or compositional visual concepts, resulting in edits that deviate from the intended modification.
To overcome the inherent ambiguity of text, recent works~\cite{Barvisual,wang2023images,yang2023imagebrush,chen2025edit,li2025visualcloze,gong2025relationadapter,lu2025pairedit}
have turned to \emph{visual prompts}—paired examples consisting of a source and a target image that explicitly demonstrate the desired transformation.
This paradigm, termed \textbf{\sys}~\cite{chen2025edit}, requires the model to faithfully apply the transformation illustrated in the visual prompt $(A, A')$ to a new query image $B$, as shown in~\figref{edittransfer}.
Compared with textual instructions, visual prompts provide concrete and unambiguous guidance, making edit transfer a promising direction for controllable and faithful image editing.

\begin{figure*}[!t]
    \includegraphics[width=\linewidth]{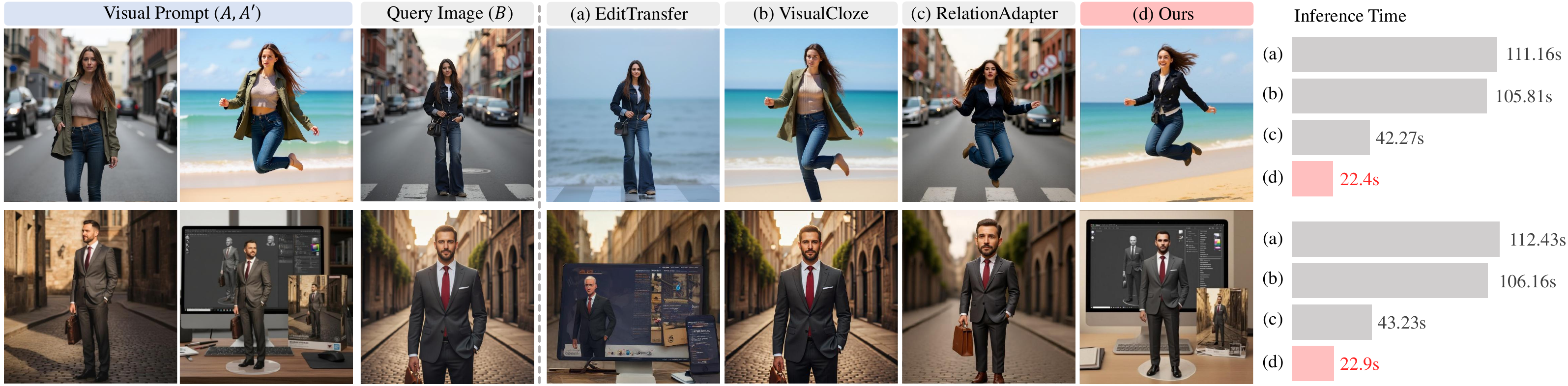}
    \vspace{-7 mm}
\caption{ \textbf{Edit transfer results and inference time.}
Given a visual prompt and a query image, existing methods
(a) EditTransfer~\cite{chen2025edit}, 
(b) VisualCloze~\cite{li2025visualcloze}, and 
(c) RelationAdapter~\cite{gong2025relationadapter} often fail to faithfully reproduce the demonstrated transformation and require long per-image inference time for 1024-long-edge outputs.
Our (d) EditTransfer++ more closely follows the visual prompt while achieving much faster inference, as illustrated by the bar plots on the right.}
    \label{fig:teaser}
    \vspace{-3 mm}
\end{figure*}

%
Recent approaches~\cite{chen2025edit,li2025visualcloze,gong2025relationadapter,jiang2025personalized} increasingly build on DiT-based text-to-image (T2I) architectures~\cite{peebles2023scalable}, which offer several appealing properties: a unified tokenization scheme that permits seamless integration of additional images and the Multi-Modal Attention (MMA)~\cite{wei2020multi} mechanism that naturally supports cross-condition interactions.
These properties endow the DiT-based T2I models with in-context learning capabilities and inspire simplified edit-transfer frameworks such as VisualCloze~\cite{li2025visualcloze} and RelationAdapter~\cite{gong2025relationadapter}.
Nevertheless, \emph{faithfully reproducing the transformation demonstrated in the visual prompt remains challenging}—especially for complex compositional edits such as coupling non-rigid motion with background changes, as illustrated in~\figref{teaser}, where existing methods either miss parts of the pose change or fail to adapt the background consistently.
These failures motivate a closer examination of why T2I backbones struggle with visual-prompt-guided edit transfer.

We identify two structural mismatches between T2I backbones and the requirements of the edit transfer task:
(1) \textbf{\emph{Textual dominance.}}
T2I models are pretrained to prioritize textual conditioning, and retaining text input during fine-tuning~\cite{chen2025edit,li2025visualcloze,gong2025relationadapter} reinforces this bias.
Consequently, cross-attention continues to favor textual tokens over visual ones, causing the model to associate visual effects with specific textual cues instead of learning the transformation conveyed by the visual prompt.
(2) \textbf{\emph{Stochastic nature.}}
Diffusion-based sampling introduces inherently random denoising trajectories designed to promote output diversity.
However, edit transfer requires deterministic reproduction of a specific transformation.
Small variations in the initial noise are amplified during sampling, and the visual prompt alone cannot anchor the generation path, resulting in significant seed-induced variability and reduced adherence to the demonstrated transformation.

To address these challenges, we propose \textbf{EditTransfer++}, which refines the T2I backbone through a \emph{progressively structured training procedure} designed to reduce textual bias and stabilize sampling behavior.
We first encourage the model to rely directly on the visual prompt by removing textual conditioning during training, allowing it to learn the transformation from visual evidence alone. 
This \textbf{text-decoupled training} strengthens the influence of the visual prompt while preserving the backbone's inherent ability to incorporate text at inference when needed.
Building upon this visually grounded model, we introduce a \textbf{best--worst contrastive refinement} mechanism to further improve generation consistency. 
For each training instance, multiple outputs are sampled under different noise seeds and ranked according to their alignment with the visual prompt. 
The model is then guided to move away from the least faithful latent states and toward the most faithful ones, effectively reducing seed-induced variability and improving visual prompt adherence.
Together, this progressively refined training procedure leads to substantially improved \textbf{\emph{faithfulness}} to the demonstrated transformation and yields stable, consistent visual-prompt-guided edits.

While the progressive training procedure improves faithfulness, the overall scalability of EditTransfer++ is still limited by the in-context design, which concatenates all conditional images into a single token sequence. 
This leads to quadratic growth in computation and memory, making high-resolution generation prohibitively expensive.
To improve efficiency and broaden applicability, we incorporate a \textit{condition compression and reuse strategy} into our framework. 
This strategy reduces both sequence length and token computation by downsampling the conditional images and reusing their nearly invariant token features across inference steps. 
We systematically explore different compression configurations to identify the best balance between efficiency and performance.
With the final configuration, EditTransfer++ is capable of generating images at a long-edge resolution of $1024$ in an average of $16$ seconds.

To thoroughly evaluate the effectiveness of EditTransfer++, we construct \textbf{EditTransfer-Bench}, a comprehensive benchmark designed to measure how well a model follows the transformation demonstrated in the visual prompt. 
It covers diverse editing scenarios spanning multiple edit types and visual effects (\eg, pose changes, appearance adjustments, and style modifications), including both single-step and compositional edits.
Using this benchmark, we perform extensive quantitative and qualitative evaluations. 
Our method consistently achieves higher faithfulness to the demonstrated transformation and offers significantly improved efficiency compared with existing approaches, as illustrated in \figref{teaser}.

The main contributions are summarized as follows:
\begin{compactitem}
    \item We introduce \textbf{EditTransfer++}, a framework that substantially improves visual-prompt faithfulness while offering significantly better computational efficiency.
    \item We design a progressively structured training procedure that integrates \textit{text-decoupled training} with \textit{best--worst contrastive refinement}, effectively reducing textual bias and seed-induced variability during sampling.
    
    \item We develop \textbf{EditTransfer-Bench}, a comprehensive benchmark covering diverse editing categories (\eg, pose changes, appearance adjustments, and style modifications) across both single-step and compositional settings, and show through extensive quantitative and qualitative studies that EditTransfer++ achieves superior visual-prompt adherence and efficiency over existing methods.
\end{compactitem}

\section{Related Work}
\label{sec:related}

\begin{figure*}[!t]
    \includegraphics[width=\linewidth]{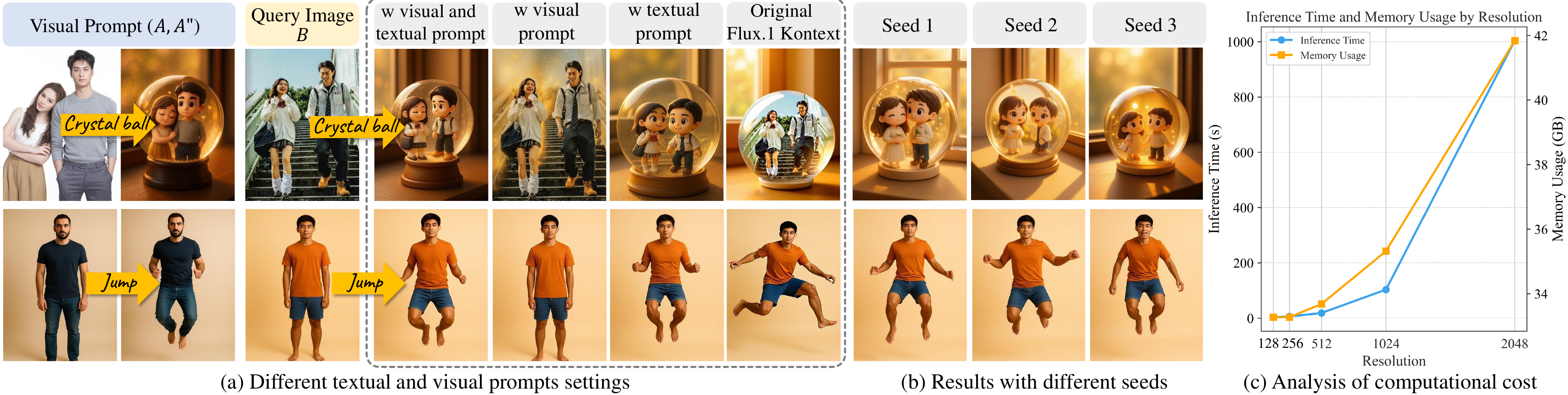}
    \vspace{-7 mm}
\caption{\textbf{Limitations of naïve DiT-based in-context design for edit transfer.} 
(a) Training with paired text–image supervision causes the model to over-associate specific visual effects with textual cues, so removing the text greatly weakens the influence of the visual prompt. 
(b) Even with the same visual prompt and text, the fine-tuned model produces divergent outputs under different random seeds, revealing low visual-prompt faithfulness.
(c) Concatenating all images into a single long token sequence makes inference time and memory usage grow rapidly with resolution, creating a major efficiency bottleneck.}
    \label{fig:motivation}
    \vspace{-2 mm}
\end{figure*}

\subsection{Diffusion-Based Text-to-Image Models}
Diffusion models such as DDPM~\cite{ho2020denoising} and DDIM~\cite{song2020denoising} have become a standard paradigm for high-quality, controllable image generation, and have been widely adopted for text-conditioned synthesis and image editing~\cite{ruiz2023dreambooth,hertz2022prompt,brooks2023instructpix2pix,101145,zhang2025stable}. 
Most diffusion-based T2I models initially adopt U-Net backbones~\cite{rombach2022high}, where spatial feature maps are progressively refined across scales. 
More recently, Transformer-based architectures such as DiT~\cite{peebles2023scalable} have gained prominence. 
DiT tokenizes an image into patch embeddings and represents it as a sequence of visual tokens processed by full self-attention. 
This \emph{unified tokenization} naturally accommodates multiple images concatenated along the token dimension, while MMA allows tokens from different conditions (\eg, text, reference images, or visual prompts) to attend to each other and exchange information.

Thanks to these properties, DiT-based models exhibit strong \emph{in-context conditioning} ability~\cite{lhhuang2024iclora}: additional visual examples can be injected as extra token sequences and treated as context during generation. 
Such capability has been leveraged in controllable generation~\cite{tan2024omini,zhang2025easycontrol,song2025omniconsistency,song2025makeanything}, instruction-guided editing~\cite{zhang2025context,huang2025photodoodle}, and, more recently, visual-prompt–guided edit transfer~\cite{chen2025edit,li2025visualcloze,gong2025relationadapter,jiang2025personalized}. 
However, because DiT backbones are pre-trained with text-dominant T2I objectives and rely on stochastic sampling, naively adapting them to edit transfer by concatenating multiple images still suffers from textual bias, seed-induced inconsistency, and substantial computational overhead, which motivates the design of our tailored training and conditioning scheme.

\subsection{Guided Image Editing with Textual and Visual Cues}

Guided image editing methods can be broadly categorized according to the type of guidance they use, most notably \emph{textual descriptions} and \emph{visual references}. 
Text-guided approaches provide flexible, high-level control through natural language, whereas visual-guided approaches supply concrete examples that capture details difficult to describe with text alone.
We briefly review both families and highlight their limitations in expressing complex transformations, which motivates our visual-prompt–guided formulation.

\Paragraph{Text-guided image editing}~\cite{hertz2022prompt,brooks2023instructpix2pix,cao_2023_masactrl,wang2024taming,avrahami2024stableflow,feng2024dit4editdiffusiontransformerimage,11223644,11223644,11218740} rely on natural language to specify the desired modification. 
Training-free strategies based on attention manipulation or injection~\cite{hertz2022prompt,cao_2023_masactrl,wang2024taming,avrahami2024stableflow,feng2024dit4editdiffusiontransformerimage} intervene in the cross- or self-attention maps of diffusion models, enabling diverse edits such as appearance changes, object replacement, and non-rigid transformations.
However, they typically require detailed prompts describing both the original scene and the intended edit, which increases prompt engineering burden and may disperse attention over irrelevant tokens, leading to coarse or incomplete edits.
To alleviate prompt complexity, instruction-based approaches~\cite{brooks2023instructpix2pix,zhang2023magicbrush,zhang2024hive} train diffusion models on large-scale datasets of image–instruction pairs, allowing users to specify edits via natural language commands.
Despite these advances, a fundamental gap remains between textual descriptions and visual content:
\emph{language alone often struggles to encode fine-grained visual semantics, such as subtle pose changes, precise spatial relations, or detailed textures.}
Consequently, even well-crafted prompts do not always yield edits that faithfully realize the intended visual transformation.

\Paragraph{Visual-guided image editing} incorporates an auxiliary image as a guidance signal to compensate for what text cannot easily express.
Early work on style transfer~\cite{Gatysstyle,Alaluf2024transfer} focuses on propagating global artistic characteristics from the guidance image to the target.
Subsequent approaches~\cite{Zhucyclegan,zhou2025attentiondistillationunifiedapproach,10418856,11212802} establish semantic correspondences between images to transfer appearance across aligned regions, while more recent methods~\cite{ChenAnyDoor,YangPaintbyExample,ChenSpecRef,he2024freeedit,chen2024mimicbrush,biswas2025PIXELS,10314461} enable localized control, such as copying hair color, clothing patterns, or object textures from the guidance image to specific target regions.
Although effective for fine-grained appearance and style transfer, these techniques largely remain limited to low- or mid-level visual attributes and generally do not model \emph{high-level transformations} such as complex non-rigid motions or action changes.

In summary, text-guided editing offers semantic flexibility but suffers from ambiguity and limited control over fine-grained details, while visual-guided editing provides precise appearance cues but is restricted in the types of edits it can express.
These limitations motivate the use of \emph{visual prompts} that explicitly demonstrate a source-to-target transformation, which we formalize as the \emph{\textbf{Edit Transfer}} task~\cite{chen2025edit} and further discuss in the next subsection.

\begin{figure*}[!t]
\centering
    \includegraphics[width=0.98\linewidth]{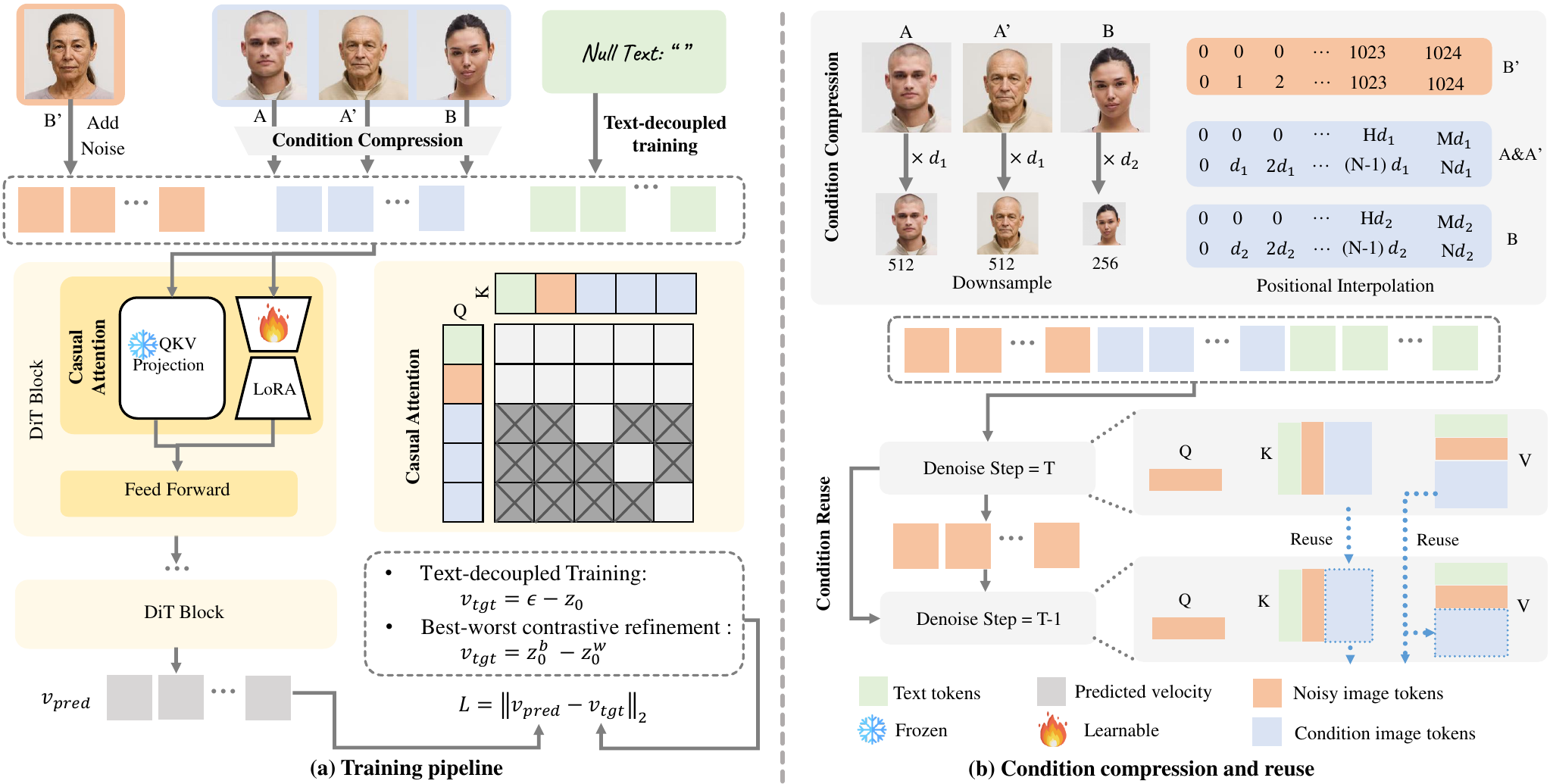}
    \vspace{-3 mm}
 \caption{\textbf{Framework of EditTransfer++.}
(a) \emph{Training pipeline.} During training, the text branch is fed with null text to enable text-decoupled learning, while the conditional images $(A, A', B)$ are downsampled for condition compression. The full token sequence $(A, A', B, B')$ is then processed by the DiT backbone, where causal attention ensures that the conditional tokens $(A, A', B)$ remain unaffected by the noisy target tokens. The network predicts the velocity of $B'$, which is used to compute the loss under the progressive training procedure (text-decoupled training followed by best--worst contrastive refinement).
(b) \emph{Condition compression and reuse.} For efficiency, we fix the output resolution to a 1024-p     ixel long edge, and apply condition compression by downsampling $(A, A')$ with ratio $d_1$ and $B$ with ratio $d_2$. To maintain spatial alignment between $B'$ and the conditional images, token positions are remapped to the original resolution according to the downsampling ratios. During inference, conditional features are computed once, cached, and reused across subsequent timesteps.}
    \label{fig:framework}
    \vspace{-2 mm}
\end{figure*}

%
\subsection{Visual-Prompt-Guided Edit Transfer} 
Inspired by the in-context learning ability of large language models (LLMs), which can learn behaviors from input–output pairs, recent works extend this idea to vision by using paired examples as \emph{visual prompts}. 
In this setting, a visual prompt consists of a source and a target image that demonstrate a specific transformation, and the goal is to transfer this transformation to a new query image—a task referred to as \emph{edit transfer}~\cite{chen2025edit}.

Early visual in-context learning methods leverage inpainting models~\cite{liu2024towards,Barvisual,Zhanggood} and masked image modeling~\cite{liu2023unifying,Wangspeak}, mainly focusing on dense prediction and low-level understanding tasks. 
ImageBrush~\cite{yang2023imagebrush} first extends this paradigm to image editing: it proposes a visual-prompt-guided framework that injects prompt features into the cross-attention layers of a U-Net-based diffusion model via an auxiliary network, thereby broadening the scope of visual in-context learning beyond analysis tasks.

Building on more powerful diffusion transformer architectures, recent works adopt a simpler token-based conditioning strategy. 
EditTransfer~\cite{chen2025edit} concatenates the visual tokens of the prompt and query images, enabling the backbone to directly attend to visual guidance during generation; with only dozens of training samples, it can adapt a pretrained text-to-image model for visual-prompt-guided editing and significantly improve complex non-rigid transformations over purely text- or reference-guided methods. 
To further enhance generalization, VisualCloze~\cite{li2025visualcloze} constructs a large-scale dataset of densely related visual tasks, where each image is annotated under multiple task formulations to encourage the learning of shared transformation patterns. 
Instead of relying solely on token concatenation as in EditTransfer and VisualCloze, RelationAdapter~\cite{gong2025relationadapter} introduces a lightweight adapter branch for visual prompt guidance and proposes a more diverse dataset covering a broader range of editing types.

Despite these advances, existing visual-prompt-guided methods still exhibit limited faithfulness to the demonstrated transformation, sensitivity to sampling randomness, and high computational cost when multiple images are concatenated into a single long token sequence. 
In contrast, our EditTransfer++ framework tackles these issues through a progressively structured training procedure and a condition compression and reuse scheme, jointly improving visual prompt adherence and inference efficiency.

\section{Methodology}
\label{sec:method}
In this section, we present the methodology of \textbf{EditTransfer++}, which enhances faithfulness and efficiency in visual-prompt-guided image editing. 
We first revisit the DiT-based T2I backbone in~\subsecref{pre} and analyze the limitations of naïve in-context learning strategies for edit transfer in~\subsecref{motivation}.
Building on these observations, we introduce a progressive training procedure (detailed in~\subsecref{tds}), comprising text-decoupled training and best–worst contrastive refinement, to gradually reduce textual bias and sampling instability.
To further enhance practical applicability, we incorporate a condition compression and reuse strategy in~\subsecref{ccs}, which reduces token length and redundant computation.

\subsection{Preliminary: DiT-based T2I model}
\label{subsec:pre}                     
DiT-based T2I models (\eg, FLUX~\cite{FLUX}) adopt token-based representations and Transformer architectures similar to LLMs, enabling in-context generation. 
In their standard design, noisy image tokens $z\in\mathbb{R}^{N\times d}$ are processed jointly with textual tokens $c_T\in\mathbb{R}^{M\times d}$ across Transformer-based DiT blocks.
Each DiT block incorporates the MMA module to fuse noisy tokens $z$ and text tokens $c_T$.
This flexible, expandable token-sequence design allows the introduction of additional visual conditions.
For example, in the TIE model FLUX.1 Kontext~\cite{labs2025flux}, the source image is encoded into visual tokens $c_V$, which are directly appended to the input sequence.
The resulting token sequence $[c_T; z; c_V]$ is then projected into query ($Q$), key ($K$), and value ($V$) matrices and processed by the MMA module to guide the edited output:
\begin{equation}
\text{MMA}([c_T; z; c_V]) = \text{softmax}\left(\frac{QK^\top}{\sqrt{d}}\right)V.
\label{attention1}  
\vspace{-0.5 mm}
\end{equation}
This bidirectional attention mechanism enables interactions among noisy tokens, visual condition tokens, and textual condition tokens, forming the foundation of DiT-based edit transfer methods.

Based on this architecture, DiT-based T2I models are typically trained under a rectified-flow objective~\cite{liu2022flow,lipman2022flow} to model a continuous transport between a noise distribution $\mathbf{z}_1\sim\pi_1$ and a data distribution $\mathbf{z}_0\sim\pi_0$.
This is achieved by parameterizing an ODE,
$\frac{d z_t}{dt}=v_\theta(z_t,t)$,
where $v_\theta$ is instantiated by the DiT network to predict the velocity of the latent path.
During training, the forward process is implemented by:
\begin{equation}
    z_t = (1-t) z_0 + tz_1, \quad t\in[0,1],
    \label{eq:forward}
\end{equation}
whose time derivative yields the ground-truth velocity:
\begin{equation}
    \frac{dz_t}{dt} = z_{1}-z_0,\quad t\in[0,1].
    \label{eq:velocity}
\end{equation}

The model is optimized to regress this velocity via:
\begin{equation}
    \theta = \arg \min_{\theta} 
    \mathbb{E}\big[\,\| (z_1 - z_0) - v_\theta(z_t, t) \|^2 \big].
    \label{eq:train}
\end{equation}                                               

While fine-tuning typically uses the same reconstruction objective as in~\eqnref{train} on a supervised dataset, the formulation is flexible and can be adapted to specific goals.
\begin{figure}[!t]
    \includegraphics[width=\columnwidth]{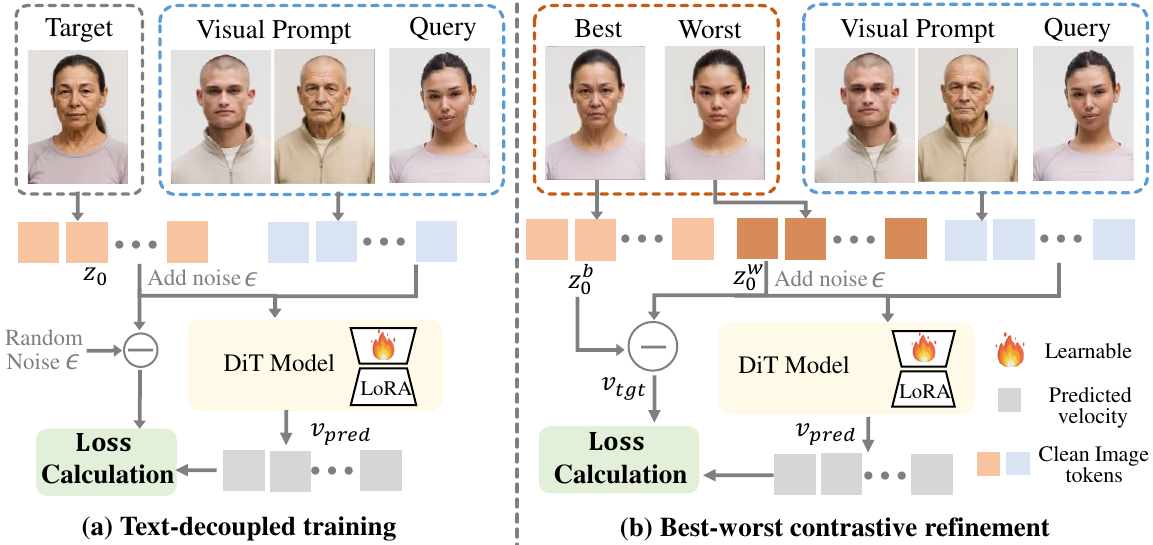}
    \vspace{-6 mm}
    \caption{ \textbf{Detailed illustration of the progressive training procedure.}
(a) In text-decoupled training, the LoRA modules are first fine-tuned using the standard velocity loss in~\eqnref{train}.
(b) In best--worst contrastive refinement, we construct a best--worst contrastive dataset and further update the LoRA with the contrastive objective in \eqnref{target} to improve generation consistency.
    }
    \label{fig:two_stage}
    \vspace{-2 mm}
\end{figure}

\begin{figure}[t]
    \includegraphics[width=\linewidth]{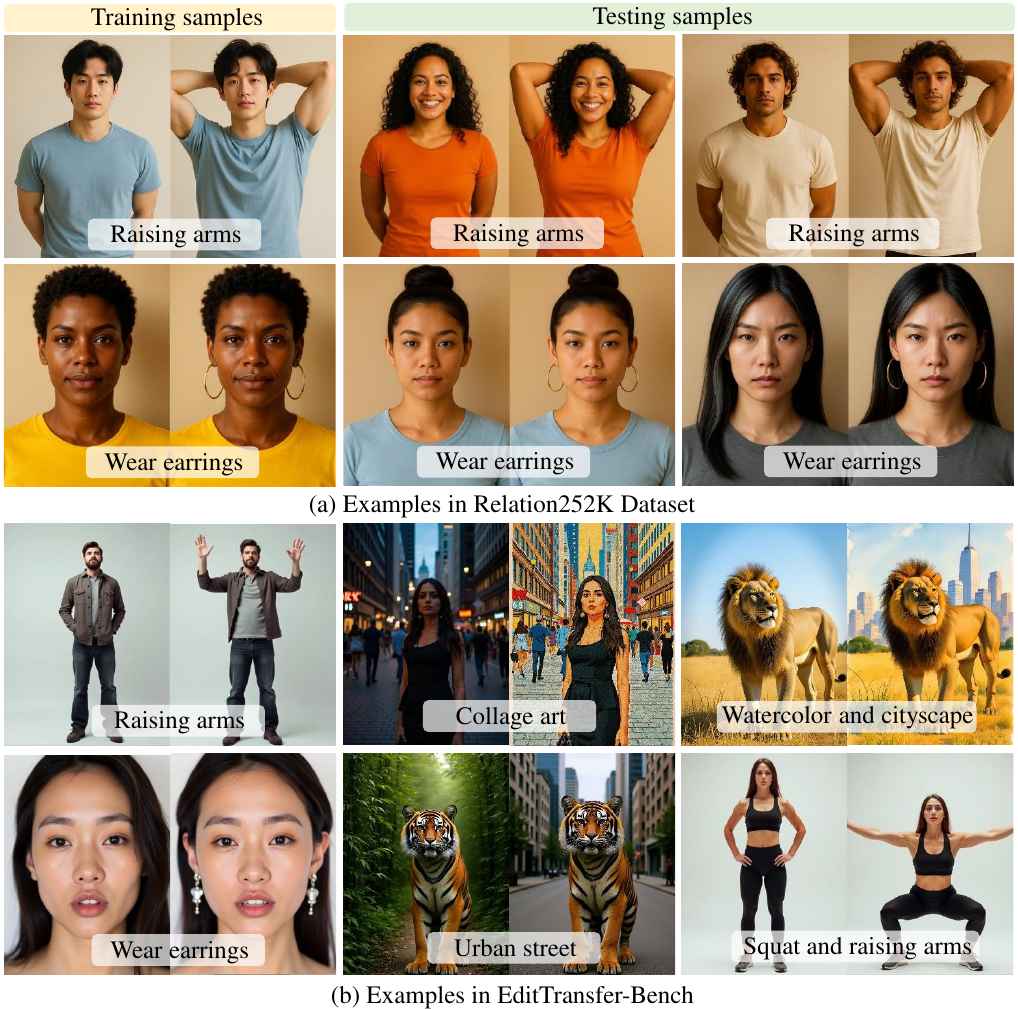  }
    \vspace{-7 mm}  
\caption{\textbf{Data samples in Relation252K and EditTransfer-Bench.}
(a) In the Relation252K~\cite{gong2025relationadapter} test split, each editing type exhibits nearly identical visual effects across samples, offering limited diversity for evaluating edit generalization.
(b) In contrast, EditTransfer-Bench introduces both diverse editing types and varied visual outcomes, enabling a more comprehensive evaluation of edit transfer capabilities.}   
    \label{fig:dataset}
    \vspace{-2 mm}
\end{figure}

\subsection{Motivation and Analysis}
\label{subsec:motivation}
To empirically examine the limitations discussed in \secref{intro}, we conduct a naïve LoRA fine-tuning experiment on FLUX.1 Kontext~\cite{labs2025flux} using standard supervised training with both text and visual prompts.
Each training sample consists of a visual prompt $(A, A')$, a query image $B$, a target image $B'$, and a textual instruction $P$.

\textbf{Observation 1 (Textual dominance).}
As shown in \figref{motivation}(a), the fine-tuned model behaves well when both text and visual prompts are provided.
However, once the textual instruction is removed, it can no longer follow the transformation indicated by $(A, A')$, while text-only editing still produces reasonable results.
This suggests that the model largely binds the visual effects to textual cues and under-utilizes the visual prompt, motivating our text-decoupled training strategy in \subsecref{tds}.

\textbf{Observation 2 (Stochastic nature).}
Even when both text and visual prompts are fixed, the model generates noticeably different outputs under different random seeds, as illustrated in \figref{motivation}(b).
Such seed sensitivity reflects the inherent stochasticity of diffusion sampling and leads to low visual-prompt faithfulness in edit transfer, which our best--worst contrastive refinement is designed to mitigate.

Beyond faithfulness limitations, the naïve in-context design also suffers from efficiency bottlenecks.
As discussed in \subsecref{pre}, DiT-based models incorporate visual guidance by concatenating all condition tokens into a single input sequence.
For edit transfer, four images—an example pair, a query image, and a target image—must be jointly encoded, so if each image is tokenized into $L$ tokens, the sequence length becomes $4L$ and the attention complexity scales as $\mathcal{O}((4L)^2)$.
The empirical measurements in \figref{motivation}(c) show that both memory usage and inference time grow rapidly with image resolution, motivating the condition compression and reuse scheme introduced in \subsecref{ccs}.

\subsection{Progressive Training Procedure}
\label{subsec:tds}
Based on the above analysis, we introduce a progressively structured training procedure, including \textit{text-decoupled training} and the \textit{best-worst contrastive refinement} to enhance visual-prompt faithfulness.

\Paragraph{Text-decoupled training.} As discussed in~\subsecref{motivation}, naively fine-tuned models tend to overfit textual instructions and exhibit weak responsiveness to visual prompts. 
To address this bias, we remove textual conditioning during training by feeding null input to the text branch (see~\figref{framework}(a)), thereby eliminating linguistic supervision.
Under this constraint, the model is forced to learn the visual relations among the visual prompt $(A, A')$, the source image $B$, and the target image $B'$. 
This simple but effective strategy enhances the influence of visual prompts while preserving the backbone’s inherent text-guided editing capability.
At inference, the model remains compatible with visual-only, text-only, or combined guidance, offering flexible control as shown in~\subsecref{discussion}.
%

\begin{figure}[!t]
    \includegraphics[width=0.98\columnwidth]{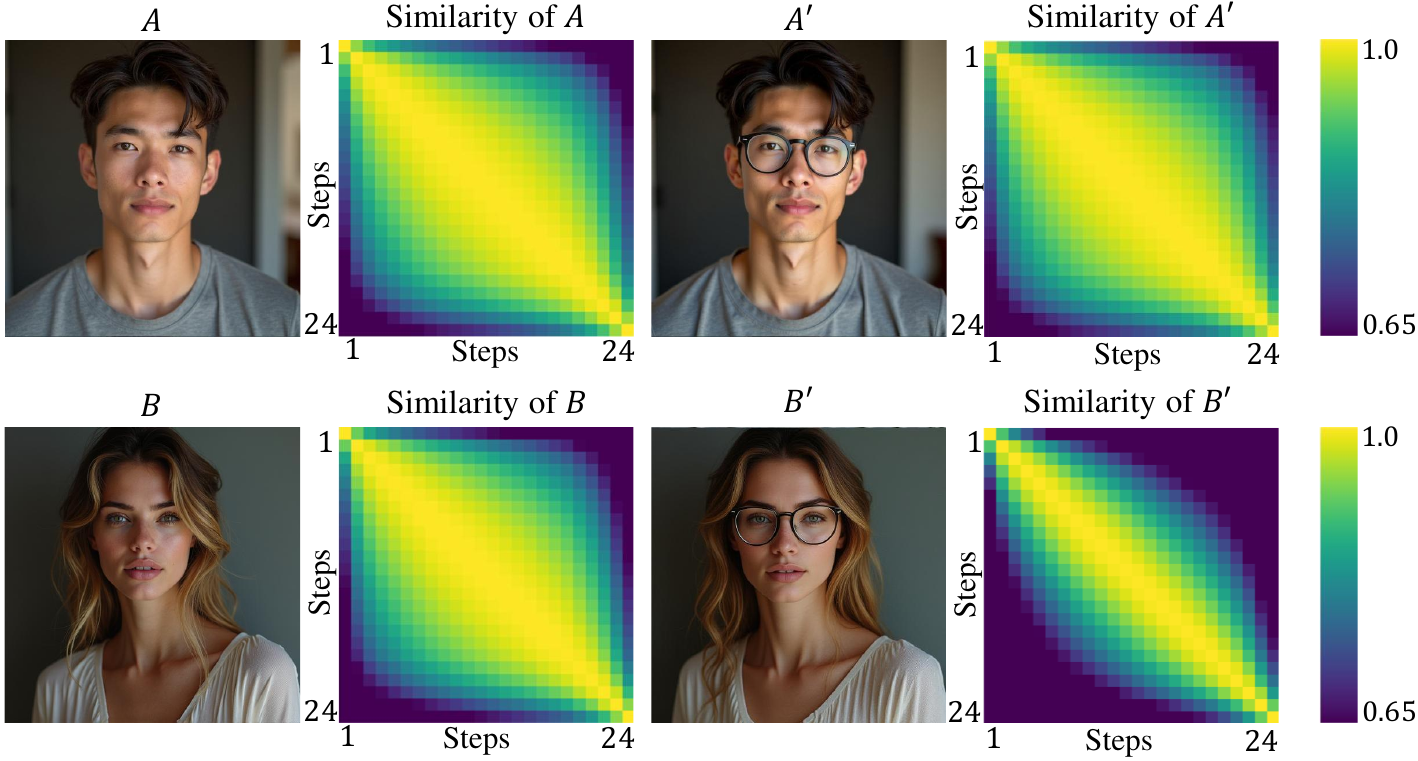}
    \vspace{-3 mm}
\caption{\textbf{Feature similarity across timesteps for each image.}
We extract intermediate features of each image $(A, A', B, B')$ at every timestep and compute pairwise similarities over time, visualized as heatmaps.
Higher similarity is shown in yellow and lower similarity in dark blue.
The conditional images $(A, A', B)$ exhibit highly stable features across timesteps, whereas the target $B'$ changes significantly, supporting our design of reusing condition features during inference.}
    \label{fig:feature_reuse}
    \vspace{-3 mm}
\end{figure}

\Paragraph{Best-worst contrastive refinement.}
Although text-decoupled training improves visual alignment, the results still vary significantly across different sampling seeds.
To mitigate this, we introduce an offline refinement stage built upon the previously fine-tuned model.
For each training sample, we generate $R$ candidate outputs using different random seeds and rank them according to the CLIP direction score~\cite{gal2022stylegan}, supplemented by manual verification.
The sample that best preserves the demonstrated transformation is designated as the \textit{best} image $I^b$, while the most deviating one is selected as the \textit{worst} image $I^w$, forming a best-worst contrastive pair.
Unlike standard fine-tuning, which regresses toward ground-truth data as in~\eqnref{train}, our refinement objective actively steers the model away from undesirable latent trajectories and toward those that align with the visual prompt.
Although the flow-matching objective in~\eqnref{train} is derived from the continuous formulation in~\eqnref{forward}, the generation process is ultimately carried out through discrete sampling.
To align with the model’s inference dynamics, we ground it in the Euler-based discretization used during sampling:
\begin{equation}
    z_{t_i-1} = z_{t_i} + (t_{i-1}-t_i) v_{\theta} (z_{t_i},t_i),
\label{eq:sampling}
\end{equation}
where the sampling process contains discrete $T$ timesteps, $t = \{t_T,...,t_0\}$ and $i\in \{T,...,1\}$.
Based on this discrete formulation, we define a contrastive velocity $v_{cts}$ that pushes $z_t$ toward the best sample $z^b$ and away from the worst sample $z^w$:
\begin{equation}
    v_{cts} = \frac{z_{t}^b-z_{t}^w}{\lambda},
    \label{eq:vtgt1}
\end{equation}
where the time-dependent term $t_{i-1} - t_i$ is replaced with the scaling constant $\lambda$.
By applying identical noise to both $z_0^b$ and $z_0^w$ under the linear interpolation in~\eqnref{forward}, \eqnref{vtgt1} simplifies to:
\begin{equation}
    v_{cts} = \frac{z_{0}^b-z_{0}^w}{\lambda}.
    \label{vtgt2}
\end{equation}

The refinement stage then minimizes the discrepancy between the model velocity and this contrastive target:
\begin{equation}
    \arg \min_{\theta} 
    \mathbb{E}\big[\,\| (\frac{z_{0}^b-z_{0}^w}{\lambda}) - v_\theta(z_t, t) \|^2 \big],
    \label{eq:target}
\end{equation}
where $\theta$ denotes the same LoRA parameters optimized during the text-decoupled training.
After this contrastive refinement, the model exhibits improved faithfulness to visual prompts.
This overall progressive training procedure is illustrated in~\figref{two_stage}.

\subsection{Condition Compression and Reuse}
\label{subsec:ccs}
To improve efficiency and broaden applicability, we integrate a condition compression and reuse strategy, effectively reducing both the sequence length and the attention computation cost.

\begin{table}[!t]
\centering
    \caption{\textbf{Quantitative comparisons of edit transfer methods.} Bold indicates the best result.}
    \begin{tabular}{c@{\hspace{0.2cm}}c@{\hspace{0.3cm}}c@{\hspace{0.4cm}}c@{\hspace{0.4cm}}c@{\hspace{0.2cm}}c@{\hspace{0.2cm}}c}
    \toprule 
    \multirow{2}{*}{\textbf{Method}} & \multicolumn{2}{c}{\textbf{Fidelity}} & \multicolumn{2}{c}{\textbf{Alignment}} & \multicolumn{2}{c}{\textbf{Consistency}} \\
    \cmidrule(r){2-3} \cmidrule(r){4-5} \cmidrule(r){6-7} 
    & DS$\downarrow$ & GPT-F$\uparrow$ &  CDS$\uparrow$  &GPT-A$\uparrow$ & SR$\uparrow$ & Var$\downarrow$ \\
    \midrule
    \rowcolor[HTML]{EFEFEF} 
    \multicolumn{7}{c}{Relation252K: Image Editing} \\
    \midrule
    ET~\cite{chen2025edit} & 0.202 &6.751 &0.318 &7.209 &0.672 &3.39e-3\\
    VC~\cite{li2025visualcloze} &0.217 &7.105 &0.236 &6.156 & 0.155 &4.17e-3 \\
    RA~\cite{gong2025relationadapter} &\textbf{0.177} &\textbf{9.128}  &0.363 & 8.991 &0.889 &1.12e-3 \\
    Ours &0.259 &8.911  &\textbf{0.401} &\textbf{9.041} &\textbf{0.949} &\textbf{1.11e-3} \\
    \midrule  
    \rowcolor[HTML]{EFEFEF} 
    \multicolumn{7}{c}{Relation252K: Low-Level} \\
    \midrule
    ET~\cite{chen2025edit} &0.212 &5.883 &0.189 &4.202 &0.167 &2.90e-3 \\
    VC~\cite{li2025visualcloze} &0.253 &6.824&\textbf{0.333} &6.201  & 0.354 &1.39e-3 \\
    RA~\cite{gong2025relationadapter} &\textbf{0.202} &\textbf{7.158} &0.257 & 5.753& \textbf{0.451}&8.12e-4 \\
    Ours & 0.260 & 5.992 & 0.245&\textbf{6.889} & 0.437&\textbf{7.26e-4} \\
    \midrule
    \rowcolor[HTML]{EFEFEF} 
    \multicolumn{7}{c}{Relation252K: Customized Generation} \\
    \midrule
    ET~\cite{chen2025edit} &\textbf{0.232} &6.200 &0.298 &6.812 &0.401 &4.09e-3 \\
    VC~\cite{li2025visualcloze} &0.297 &6.713 &0.264 & 7.489& 0.378 &3.81e-3 \\
    RA~\cite{gong2025relationadapter} &0.279 &8.505 &0.384 & 8.495&0.739 &1.27e-3 \\
    Ours &0.337 &\textbf{8.670} &\textbf{0.389} &\textbf{8.826} &\textbf{0.744} &\textbf{1.20e-3} \\    
    \midrule
    \rowcolor[HTML]{EFEFEF} 
    \multicolumn{7}{c}{EditTransfer-Bench} \\
    \midrule
    ET~\cite{chen2025edit} &\textbf{0.157}&6.454 &0.148 &6.122 &0.230 &5.70e-3  \\
    VC~\cite{li2025visualcloze} &0.159 & 7.243&0.240  & 6.322&0.322&4.40e-3 \\
    RA~\cite{gong2025relationadapter} &0.159 &\textbf{8.925} &0.229 &7.580 &0.347 &3.68e-3 \\
    Ours&0.169 &8.631  &\textbf{0.292} &\textbf{8.152} &\textbf{0.397}&\textbf{2.37e-3} \\    
    \bottomrule 
    \end{tabular}
    \vspace{-3 mm}
    \label{tab:quantitative}
\end{table}
\begin{table}[!t]
\centering
    \caption{\textbf{Comparison of inference time and memory consumption across methods.} }
    \vspace{-1 mm}
    \begin{tabular}{>{\centering\arraybackslash}p{1.3cm}>{\centering\arraybackslash}p{0.9cm}>{\centering\arraybackslash}p{0.9cm}>{\centering\arraybackslash}p{0.9cm}>{\centering\arraybackslash}p{0.9cm}>{\centering\arraybackslash}p{0.9cm}}
    \toprule 
    \textbf{Method} & \textbf{ET}~\cite{chen2025edit} & \textbf{VC}~\cite{li2025visualcloze}  & \textbf{RA}~\cite{gong2025relationadapter}& \textbf{PE}~\cite{lu2025pairedit} & \textbf{Ours}\\
    \midrule
     Time(s) &75.05 & 70.91 &29.93 &28.19 &\textbf{15.93}  \\
     Memory(GB)&39.84 &37.81  & 40.54& \textbf{33.95} &41.55  \\
    \bottomrule
    \end{tabular}
    \vspace{-3 mm}
    \label{tab:efficiency}
\end{table}
\Paragraph{Condition compression.} 
Let the conditional images $(A, A', B)$ have resolution $H \times W$, and the target output $B'$ be $M \times N$.
We downsample the visual prompt $(A, A')$ to $\frac{M}{2} \times \frac{N}{2}$ and the query image $B$ to $\frac{M}{4} \times \frac{N}{4}$.
The corresponding downsampling ratios $d_1$ for $(A, A')$ and $d_2$ for $B$ are:
\begin{equation}
    d_1=\frac{H}{M/2},\quad d_2=\frac{H}{M/4}.
\end{equation}

We justify this choice in~\subsecref{AS} through an empirical study of different configurations, identifying the best trade-off between efficiency and editing quality.  
This simple adjustment reduces the sequence length by nearly $61\%$ while preserving editing quality.
To maintain spatial correspondence among $(A, A', B, B')$ after compression, we apply positional interpolation.  
For a token at position $(i, j)$ in the resized conditional image, its original position $(P_i, P_j)$ is computed as:
\begin{equation}
    P_i=i\times d_*,\quad P_j=j\times d_*,\quad d_* \in \{d_1, d_2\}.
\end{equation}
This mapping preserves alignment within the compressed token sequence.

\Paragraph{Condition reuse.}
While the conditional images $(A, A', B)$ remain clean throughout the denoising process, the naive fine-tuning pipeline re-encodes them at every timestep, introducing redundant computation.
As shown in~\figref{feature_reuse}, the pairwise feature similarity across timesteps is consistently high for $(A, A', B)$, whereas it changes for $B'$, indicating that conditional features are temporally stable.
To exploit this, we adopt causal attention to further isolate the interference from evolving noisy and text tokens.
During attention calculation, $(A, A', B)$ tokens are restricted to attend only to themselves, as illustrated in~\figref{framework}(a).
The casual attention enables a lightweight KV-cache mechanism during inference: condition features of $(A, A', B)$ are computed once and reused across timesteps, as shown in~\figref{framework}(c).
The design reduces inference time significantly with minimal memory overhead, enhancing the practicality of EditTransfer++ for high-resolution generation.

\begin{figure}[t]
    \includegraphics[width=\linewidth]{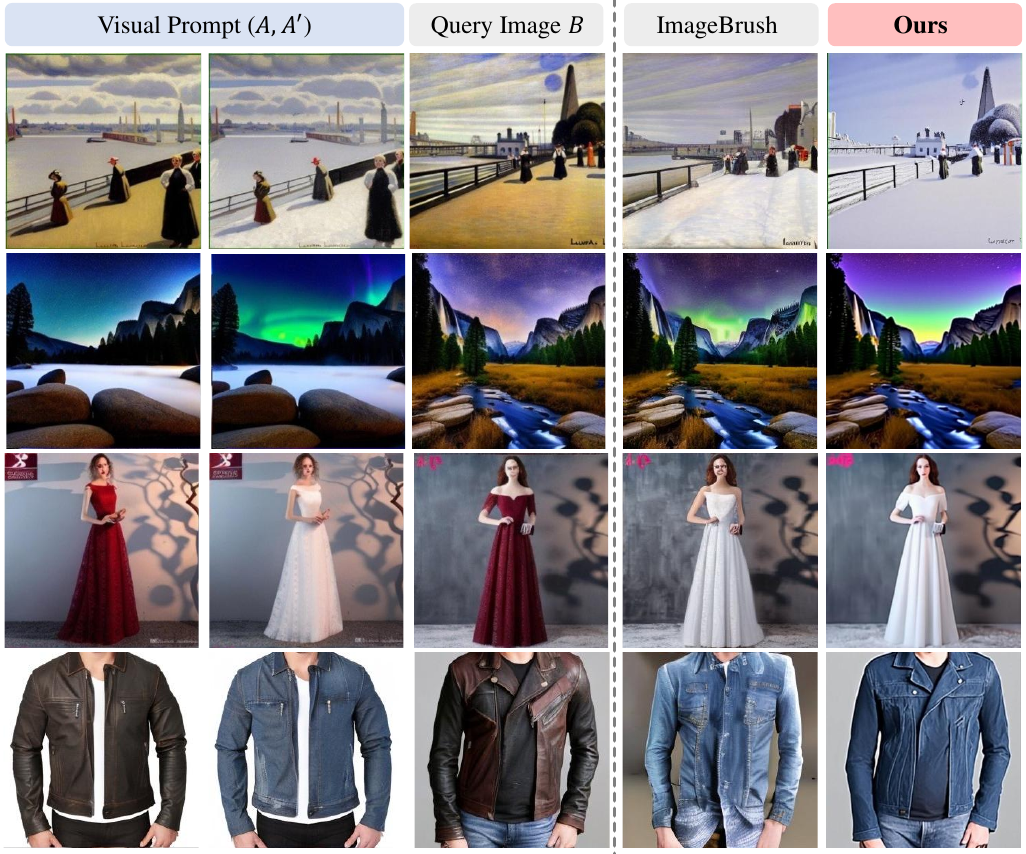}
    \vspace{-7 mm}  
\caption{\textbf{Qualitative comparisons with ImageBrush~\cite{yang2023imagebrush}.}
Since ImageBrush is not open-source, we use the examples provided in its original paper for comparison.
Given the same visual prompts and query images, our method not only follows the demonstrated transformation more faithfully, but also better preserves the identity and structure of the query image.}
    \label{fig:imagebrush}
    \vspace{-2 mm}
\end{figure}
\begin{figure}[t]
    \includegraphics[width=\linewidth]{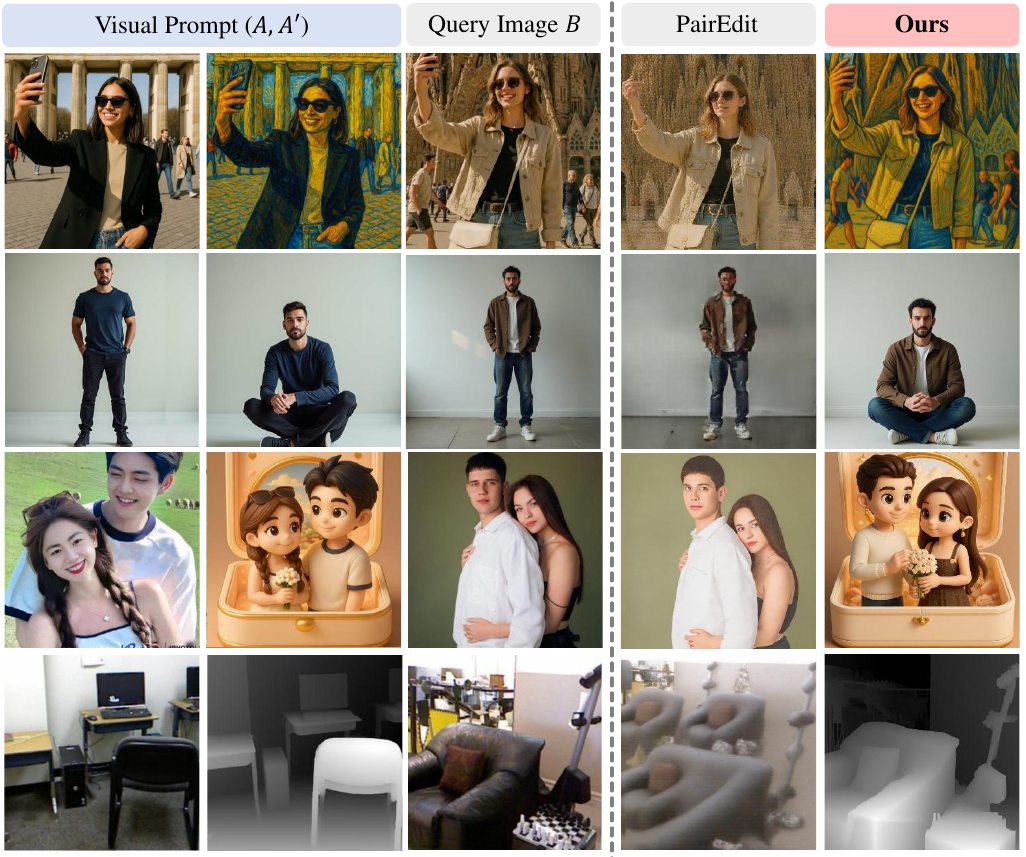}
    \vspace{-7 mm}  
\caption{\textbf{Qualitative comparisons with PairEdit~\cite{lu2025pairedit}.}
We compare our method with PairEdit on image editing, customization, and low-level understanding tasks.
Although PairEdit requires training a separate LoRA for each semantic variation, it often produces only subtle changes, whereas our method achieves stronger transformations that better follow the visual prompt.
} 
    \label{fig:pairedit}
    \vspace{-2 mm}
\end{figure}

\begin{figure*}[!t]  
\centering
    \includegraphics[width=0.95\linewidth]{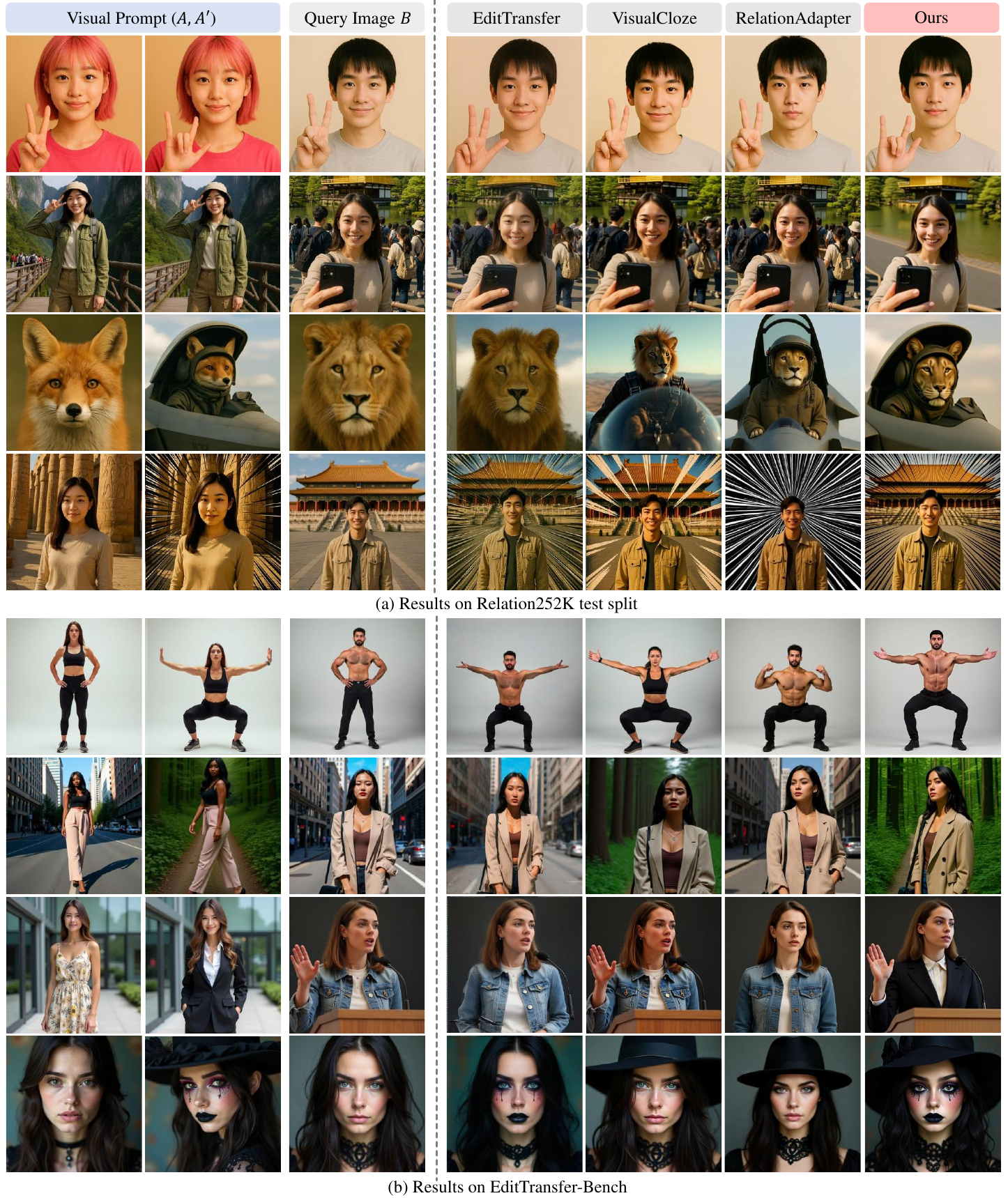  }
    \vspace{-3 mm}
\caption{\textbf{Qualitative comparisons with baselines on Relation252K and EditTransfer-Bench.}
For each example, given a visual prompt $(A,A')$ and a query image $B$, we compare the results of
EditTransfer~\cite{chen2025edit}, VisualCloze~\cite{li2025visualcloze}, RelationAdapter~\cite{gong2025relationadapter}, and our EditTransfer++.
(a) Results on the Relation252K test split~\cite{gong2025relationadapter}.
(b) Results on the proposed EditTransfer-Bench.
Across diverse editing types and difficulty levels, our method more faithfully transfers the demonstrated transformation while better preserving the query image content.}
    \label{fig:baselines}
    \vspace{-3 mm}
\end{figure*}

\section{Experiments}
\label{sec:Experiments}

\subsection{Implementation Details}
We initialize our model from FLUX.1 Kontext~\cite{labs2025flux}.
Following the 3D RoPE embedding in FLUX.1 Kontext, each conditional image is assigned a constant offset to separate context tokens from target tokens.
Specifically, token positions are represented as triplets $(k,i,j)$, where $(0,i,j)$ is used for the target $B'$, $(1,i,j)$ for the query image $B$, and $(2,i,j)$ for the visual prompt $(A,A')$, with $(i,j)$ denoting spatial coordinates.
The scaling constant in~\eqnref{target} is set to $0.2$.
We fine-tune the model using LoRA with rank $128$ and a learning rate of $1\text{e-}4$.
Training is conducted on two H20 GPUs with an accumulated batch size of $4$, using the AdamW optimizer~\cite{loshchilov2017decoupled} and bfloat16 mixed precision.
The longer side of the target image $B'$ is fixed at $1024$ pixels.
We train for $125{,}000$ iterations on the training split of the Relation252K dataset~\cite{gong2025relationadapter}.

\begin{figure}[!t]
\centering
    \includegraphics[width=0.91\linewidth]{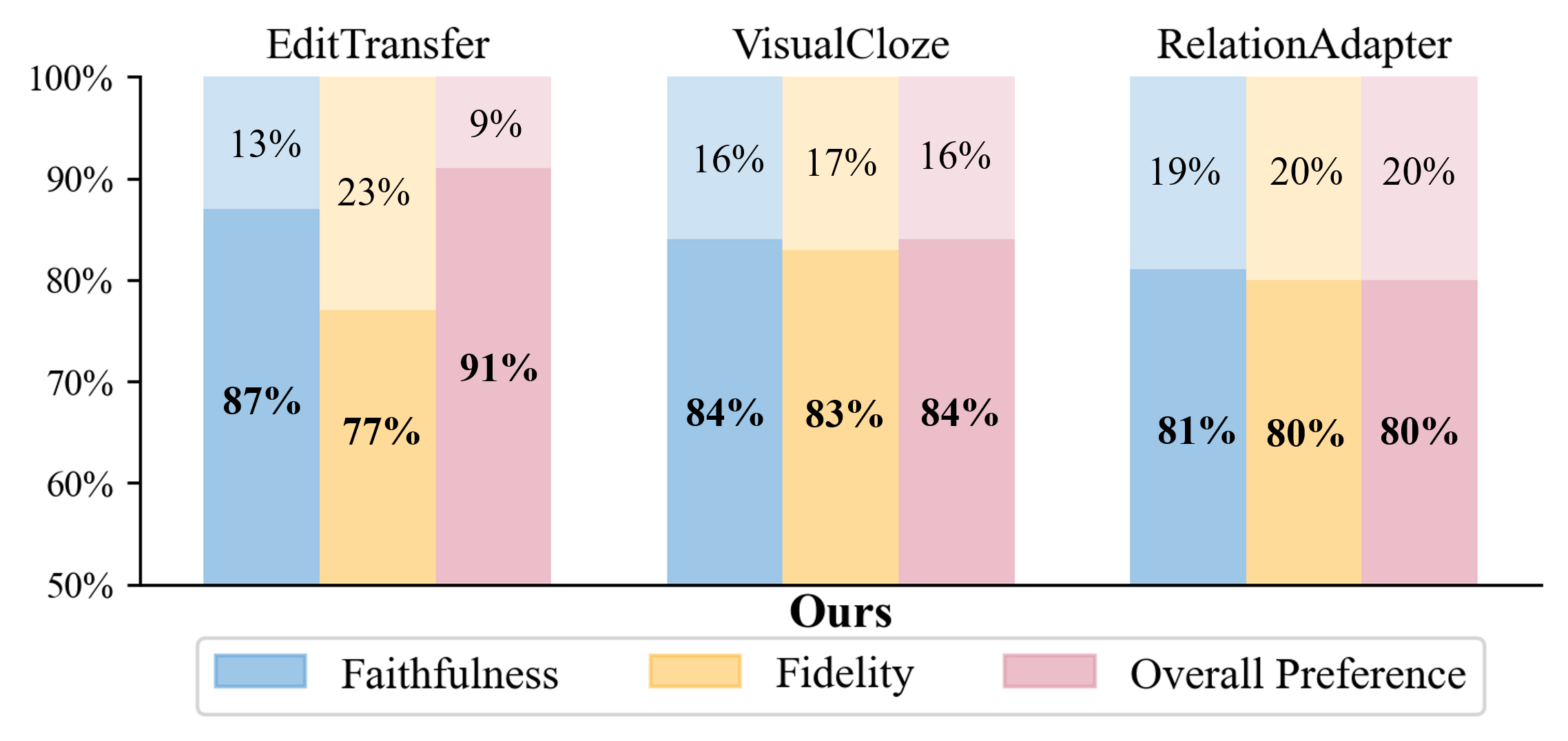  }
    \vspace{-3 mm}
\caption{\textbf{User study results.}
Bars report the proportion of participants who preferred EditTransfer++ over each baseline
(EditTransfer~\cite{chen2025edit}, VisualCloze~\cite{li2025visualcloze}, RelationAdapter~\cite{gong2025relationadapter}) in terms of faithfulness, fidelity, and overall preference.
Across all three criteria, the majority of users favor our method.}
    \label{fig:user_study}
    \vspace{-3 mm}
\end{figure}

\subsection{Benchmarks}
We evaluate our method on both the test split of Relation252K~\cite{gong2025relationadapter} and our constructed EditTransfer-Bench.
The Relation252K~\cite{gong2025relationadapter} test split spans image editing, image customization, and low-level understanding.
Its editing subset, however, exhibits limited diversity, with each editing type appearing in nearly fixed visual forms across the dataset.
For example, as shown in~\figref{dataset}(a), the ``raising arms'' edits remain identical throughout the dataset.   
This reduces the benchmark’s capacity to faithfully assess edit transfer generalization, particularly under diverse visual realizations of the same editing type.
To address this limitation, we construct EditTransfer-Bench, a benchmark focused on image editing and designed to systematically evaluate model performance on diverse edit transfer tasks.
Diversity is reflected in two aspects:
(1) a wide variety of editing types, and
(2) multiple distinct visual realizations within each type.
As illustrated in~\figref{dataset}(b), EditTransfer-Bench consists of four primary editing categories, covering both single-edit and compositional-edit scenarios:
\begin{compactitem}
    \item \textbf{Non-rigid editing} focuses on complex motion patterns with high variability.
    \item \textbf{Style transfer} targets uncommon styles constructed using community-released LoRAs.
    \item \textbf{Background change} applies edits primarily to the background rather than the main subjects, a setting rarely seen in the training data.
    \item \textbf{Appearance transfer} emphasizes diverse local appearance modifications, such as adding glasses or changing clothing color.
\end{compactitem}
These four categories and their combinations form a systematic benchmark to evaluate the model’s generalization.

\subsection{Evaluation Metrics}
We quantitatively evaluate our proposed method against baseline models using automatic metrics, large vision-language model (VLM) scores, and human evaluations.

\Paragraph{Automatic metrics.}
Our automatic evaluation covers three aspects of the edit transfer behavior:
\emph{editing fidelity to the query image}, \emph{visual-prompt faithfulness}, and \emph{inference efficiency}.
\textit{(1) Editing fidelity to the query image.}
Since edit transfer is instantiated as an image editing task, a reasonable edit should preserve content that is not intended to change (\eg, identity, global structure, and background context).
To measure this, we compute the DINO-ViT self-similarity distance (DS)~\cite{tumanyan2022splicing} between the query image $B$ and the generated output $B'$.
%
\textit{(2) Visual-prompt faithfulness.}
We characterize faithfulness to the visual prompt along two dimensions:
\textbf{\emph{(a) Alignment with the visual prompt.}}
To evaluate whether the applied edit matches the transformation demonstrated by the visual prompt, we follow CLIP directional similarity~\cite{gal2022stylegan} and compute the CLIP Direction Score (CDS) using CLIP ViT-L/14, which measures how well the editing direction of $(B,B')$ aligns with that of $(A,A')$.
\textbf{\emph{(b) Consistency across seeds.}}
Beyond aligning with the visual prompt on average, a faithful edit transfer model should reproduce the same transformation consistently under different random seeds.
To assess this consistency, we measure the variance of CDS across generations (Var), where lower values indicate more stable behavior with respect to the intended editing direction.
\textit{(3) Inference efficiency.}
For efficiency, we report inference time and peak GPU memory usage for each method.

\Paragraph{VLM scores.} To further assess editing quality from a human-centered perspective, we employ the VLM GPT-4o~\cite{openai2024gpt4technicalreport} to provide three metrics:
\textit{(1) GPT-F}, assessing the fidelity between the query image $B$ and the output $B'$.
\textit{(2) GPT-A}, evaluating the alignment of the intended transformation between the visual prompt $(A, A')$ and $(B, B')$; and 
\textit{(3) Success Rate}, estimating how consistently the model applies the intended transformation across different random seeds.
Following the protocol in RelationAdapter~\cite{gong2025relationadapter}, GPT-F and GPT-A are scored on a $0$–$10$ scale, where higher scores indicate better performance.
The success decision is binary, and we report the average success rate across multiple generations.

\Paragraph{User study.} To further evaluate the effectiveness of our method, we conduct a user study comparing it with existing approaches.
In each questionnaire, participants are shown the visual prompt $(A, A')$, the query image $B$, and two output images: one generated by EditTransfer++ and the other by a baseline method.
%
%
Participants are asked to answer the following questions:
\begin{compactitem}
    \item \textbf{Fidelity with the query image:} Which result better preserves consistency with the source image $B$?
    \item \textbf{Alignment with visual prompt:} Which result better aligns with the demonstrated transformation from $A$ to $A'$?
    \item \textbf{Overall Quality:} Which image do you prefer overall?
\end{compactitem}

\begin{table}[!t]
\footnotesize
\renewcommand{\arraystretch}{1.2}
    \caption{\textbf{Quantitative results of ablation study.} }
    \begin{tabular}{>{\centering\arraybackslash}p{2.2cm}@{\hspace{0.2cm}}>{\centering\arraybackslash}p{0.8cm}@{\hspace{0.2cm}}>{\centering\arraybackslash}p{0.8cm}@{\hspace{0.2cm}}>{\centering\arraybackslash}p{0.9cm}@{\hspace{0.2cm}}>{\centering\arraybackslash}p{1.3cm}@{\hspace{0.2cm}}>{\centering\arraybackslash}p{1.3cm}}
    \toprule 
    \textbf{Method} & \textbf{DS}$\downarrow$ & \textbf{CDS}$\uparrow$& \textbf{Var}$\downarrow$ &  \textbf{Time(s)}$\downarrow$ & \textbf{Mem(GB)}$\downarrow$ \\
    \midrule
    \rowcolor[HTML]{EFEFEF} 
    \multicolumn{6}{c}{\textbf{Relation252K}} \\
    \midrule
    $M_{b}$: Base model   & 0.246 & 0.370  &9.10e-3 &72.14 &39.00 \\
    $M_{1}$: $M_{b}+$ TD  &\textbf{0.232} & 0.374  &3.54e-3     &72.14  &39.00\\
    $M_{2}$: $M_{1}+$ RF &0.247 &\textbf{0.379}  &2.43e-3 &22.56  &\textbf{38.29}\\
    $M_{3}$: $M_{2}+$ DS &0.249 &0.375  &1.09e-3 &\textbf{16.71}  &43.23 \\
    $M_{4}$: $M_{3}+$ RU &0.268  &0.376  &\textbf{1.02e-3} &\textbf{16.71} &43.23\\
    \midrule
    \rowcolor[HTML]{EFEFEF} 
    \multicolumn{6}{c}{\textbf{EditTransfer-Bench}} \\    
    \midrule
    $M_{b}$: Base model   &0.171 &0.287  &3.02e-3 &67.04 &37.62\\
    $M_{1}$: $M_{b}+$ TD  &0.184 &\textbf{0.311}  &3.81e-3 &17.39 &61.07\\
    $M_{2}$: $M_{1}+$ RF &\textbf{0.167} &0.304  &3.07e-3 &20.60 &\textbf{36.68}\\
    $M_{3}$: $M_{2}+$ DS &0.173 &0.288  &\textbf{2.17e-3}&\textbf{15.15} &39.88 \\
    $M_{4}$: $M_{3}+$ RU &0.169 &0.292 &2.37e-3 &\textbf{15.15} &39.88\\
    \bottomrule
    \end{tabular}
    \vspace{-1 mm}
    \label{tab:ablation}
\end{table}
\subsection{Comparisons with State-of-the-art Methods}
\Paragraph{Baselines.}
We compare EditTransfer++ with representative visual prompt–guided editing methods and a related pairwise editing method:

\begin{compactitem}
    \item \textbf{U-Net–based visual prompt editing.}
    ImageBrush~\cite{yang2023imagebrush} is a U-Net-based method that first broadened visual in-context learning from low-level understanding tasks to image editing. 
    It injects visual-prompt features into the cross-attention layers of a diffusion U-Net.
    Since its implementation and training code are not publicly released, we refer to the qualitative results reported in the original paper for comparison.
    \item \textbf{DiT-based edit transfer methods.}
    EditTransfer (ET)~\cite{chen2025edit} is the first work to exploit the in-context learning ability of DiT-based models for edit transfer, primarily targeting non-rigid edits. 
    Because the original model is trained on a small-scale dataset, we retrain it on Relation252K~\cite{gong2025relationadapter} for a fair comparison.
    VisualCloze (VC)~\cite{li2025visualcloze} formulates edit transfer as a masked prediction/inpainting problem on a large-scale multi-task dataset.
    RelationAdapter (RA)~\cite{gong2025relationadapter} introduces a lightweight adapter branch to fuse visual prompts with DiT features.
    \item \textbf{Pairwise editing with learned variations.}
    PairEdit (PE)~\cite{lu2025pairedit} learns a semantic variation from a set of source–target pairs and applies it to new inputs.
    This setting is different from our visual-prompt–guided edit transfer, which directly follows the specific transformation shown in a single example pair.
    Therefore, we include PE only in qualitative comparisons.
\end{compactitem}
%

\begin{table}[!t]
\footnotesize
\renewcommand{\arraystretch}{1.2}
    \caption{\textbf{Ablation study on the downsampling ratios.}}
    \begin{tabular}{>{\centering\arraybackslash}p{1.5cm}@{\hspace{0.1cm}}>{\centering\arraybackslash}p{1.1cm}@{\hspace{0.2cm}}>{\centering\arraybackslash}p{1.1cm}@{\hspace{0.2cm}}>{\centering\arraybackslash}p{1.1cm}@{\hspace{0.2cm}}>{\centering\arraybackslash}p{1.1cm}@{\hspace{0.2cm}}>{\centering\arraybackslash}p{1.1cm}}
    \toprule 
    \textbf{(A,A')} & \textbf{1024} & \textbf{512} & \textbf{256} & \textbf{512} & \textbf{256} \\
    \textbf{(B)} & \textbf{1024} & \textbf{512} & \textbf{512} & \textbf{256} & \textbf{256} \\
    \midrule
    \rowcolor[HTML]{EFEFEF} 
    \multicolumn{6}{c}{\textbf{Relation252K}} \\
    \midrule
    DS $\downarrow$ &\textbf{0.232} &0.250 & 0.246&0.247 &0.246 \\
    CDS $\uparrow$ &\textbf{0.363} &0.362 &0.340 &0.360 &0.350 \\
    Time(s) $\downarrow$ &72.14 &26.02 &19.88 &22.56&\textbf{17.45}\\
    Memory(GB)$\downarrow$ &39.00 &38.38&38.25 & 38.29 &\textbf{37.64}\\
    \midrule
    \rowcolor[HTML]{EFEFEF} 
    \multicolumn{6}{c}{\textbf{EditTransfer-Bench}} \\
    \midrule
    DS $\downarrow$ &\textbf{0.165} &0.171 &0.172 & 0.173 &0.175 \\
    CDS $\uparrow$ &0.240 &\textbf{0.304} &0.295 &0.298 &0.291 \\
    Time(s) $\downarrow$ &67.04&29.74  &18.86&20.60 &\textbf{15.50}\\
    Memory(GB)$\downarrow$ &37.62 &36.90&36.64 &36.68 &\textbf{36.66} \\
    \bottomrule 
    \end{tabular}
   
    \vspace{-1 mm}
    \label{tab:downsample}
\end{table}

\Paragraph{Editing fidelity to the query image.}
We first analyze how well each method preserves content in the query image $B$ that is not meant to change.
As shown in the first row of~\figref{baselines}(b), VisualCloze~\cite{li2025visualcloze} tends to overfit the visual prompt and fails to maintain subject identity in $B$, producing an edited image $B'$ whose global appearance drifts toward the prompt.
Similarly, in the fourth row of~\figref{imagebrush}, ImageBrush~\cite{yang2023imagebrush} struggles to preserve fine-grained jacket textures.
In contrast, EditTransfer++ adapts the strength of the edit according to the demonstrated transformation, preserving the subject’s shape, pose, and local appearance in $B$ when these factors are not directly involved in the edit.
Quantitatively, our method maintains competitive DS and achieves a strong VLM-based GPT-F score, while user preferences in~\figref{user_study} further indicate that edits produced by EditTransfer++ are more often judged as faithful to the query image.

\Paragraph{Visual-prompt faithfulness.}
We next evaluate how faithfully each method follows the transformation demonstrated by the visual prompt $(A,A')$, both in terms of alignment and consistency.

\textit{Alignment.}
As illustrated in the first row of~\figref{baselines}(a), EditTransfer~\cite{chen2025edit} often fails to apply the intended pose change, leaving the gesture unchanged.
In the second row, VisualCloze~\cite{li2025visualcloze} and RelationAdapter~\cite{gong2025relationadapter} partially capture the ``animal airforce'' concept, but their layouts and fine-grained appearance deviate noticeably from the visual prompt.
Moreover, \figref{baselines}(b) shows that when faced with compositional edits, existing DiT-based methods tend to complete only one of the required edits, even when full text guidance is provided.
For PairEdit~\cite{lu2025pairedit},it performs well on human facial appearance editing but struggles with low-level tasks, non-rigid edits, and customized generation, as illustrated in~\figref{pairedit}.
In contrast, EditTransfer++ can handle both seen and diverse unseen cases, including single and compositional edits, while closely matching the demonstrated transformation.
This advantage is reflected in our higher CDS and GPT-A on editing and customization benchmarks, as reported in~\tabref{quantitative}, as well as by the human preference study in~\figref{user_study}, where our method is consistently preferred for visual-prompt faithfulness.

\textit{Consistency.}
Faithfulness to a visual prompt is not only about producing one correct edit, but also about reproducing the same transformation reliably under different random seeds.
To examine this, we randomly sample $100$ test cases and generate $5$ outputs per case using a shared set of seeds for all methods.
As reported in~\tabref{quantitative}, EditTransfer++ achieves the lowest Var and the highest CLIP-based success rate (SR)     across most settings, indicating that it is more robust to sampling randomness while remaining faithful to the visual prompt.
Additional qualitative examples in the supplementary material show that, for challenging non-rigid and compositional edits, our outputs are visually more stable across seeds than those of the baselines.

\Paragraph{Inference efficiency.}
\tabref{efficiency} reports inference time and peak GPU memory usage when the long side of the generated image is set to $1024$ pixels.
EditTransfer~\cite{chen2025edit} and VisualCloze~\cite{li2025visualcloze}, which concatenate all image tokens at full resolution, incur the highest computational cost.
Although RelationAdapter~\cite{gong2025relationadapter} splits the visual prompt $(A,A')$ and the query–target pair $(B,B')$ into two streams via an adapter, it still runs relatively slowly, and the additional adapter branch increases memory usage at high resolutions.
In contrast, EditTransfer++ achieves the fastest inference with a modest memory footprint, benefiting from the proposed condition compression and reuse strategy.
This shows that our method not only improves visual-prompt faithfulness, but also offers better practical efficiency for high-resolution edit transfer.
\begin{figure}[!t]
\centering
    \includegraphics[width=\linewidth]{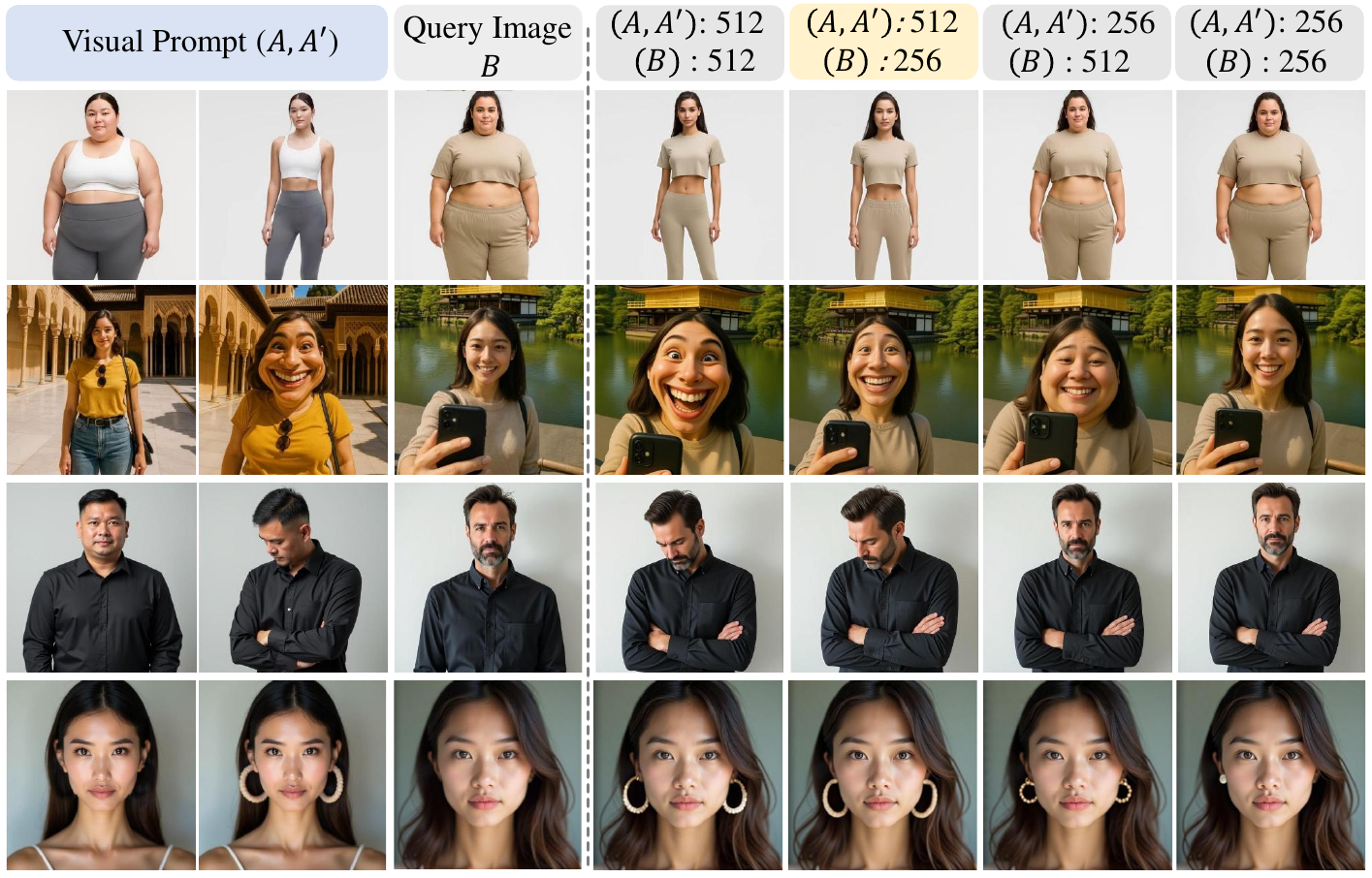  }
    \vspace{-7 mm}
\caption{\textbf{Qualitative results under different condition-compression settings.}
All outputs are generated at a 1024-pixel long edge.
Overly aggressive downsampling (both $(A,A')$ and $B$ at $256$) leads to noticeable degradation,
whereas using a $512$-pixel visual prompt $(A,A')$ and a $256$-pixel query image $B$ offers a better trade-off
between visual quality and efficiency.}
    \label{fig:resolution}
    \vspace{-3 mm}
\end{figure}

\begin{figure*}[!t]
    \includegraphics[width=\linewidth]{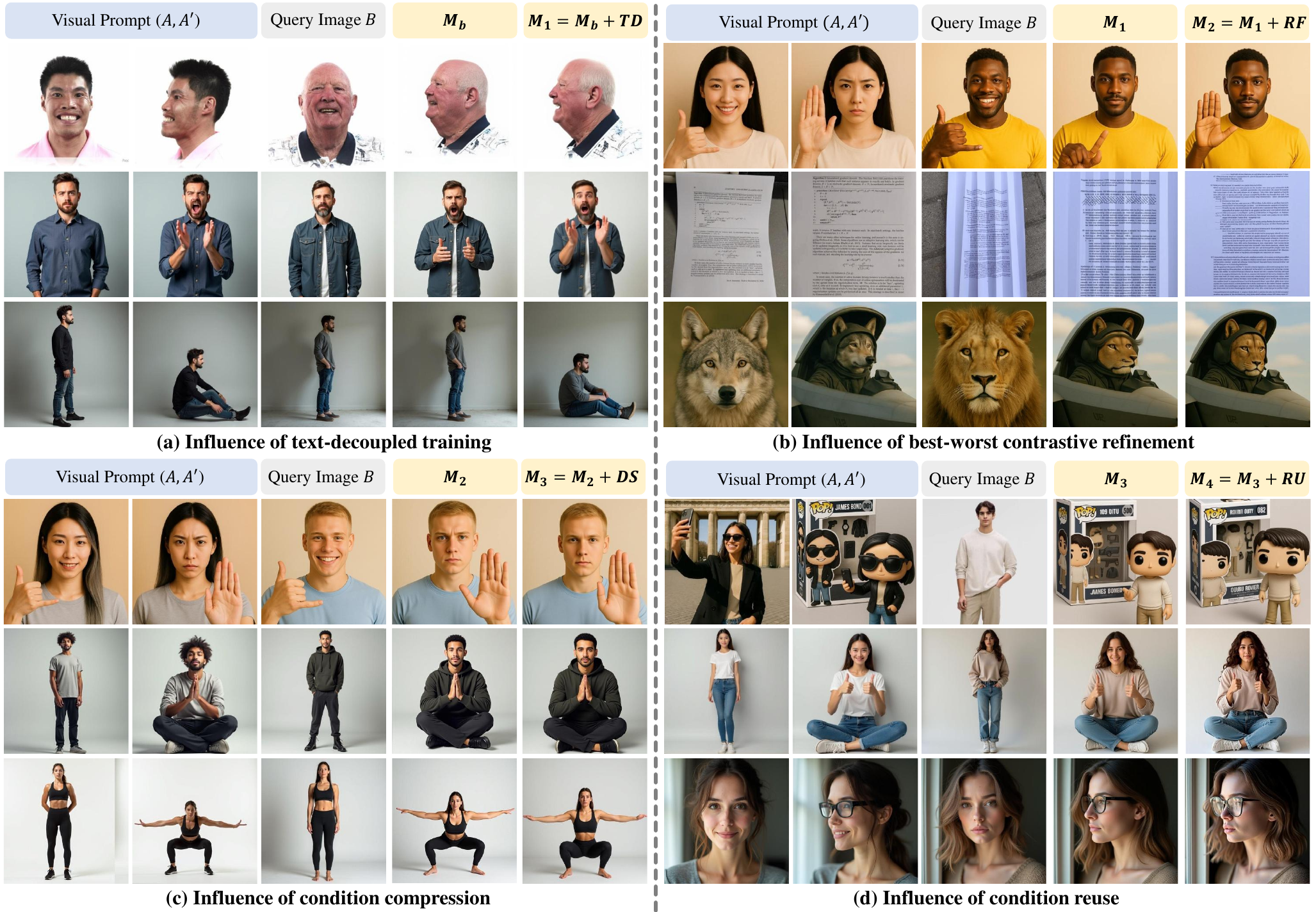}
    \vspace{-7 mm}
   \caption{\textbf{Qualitative evaluation of EditTransfer++ components.}
(a) Effect of text-decoupled training (TD); (b) effect of best--worst contrastive refinement (RF);
(c) effect of condition compression (DS); (d) effect of condition reuse (RU).
$M_b$ denotes the base model without our strategies; $M_1 = M_b+\text{TD}$,
$M_2 = M_1+\text{DS}$, $M_3 = M_2+\text{RU}$, and $M_4 = M_3+\text{RF}$ as the full EditTransfer++ model.}
    \label{fig:ablation}
    \vspace{-1 mm}
\end{figure*}
\begin{figure*}[!t]
    \includegraphics[width=\linewidth]{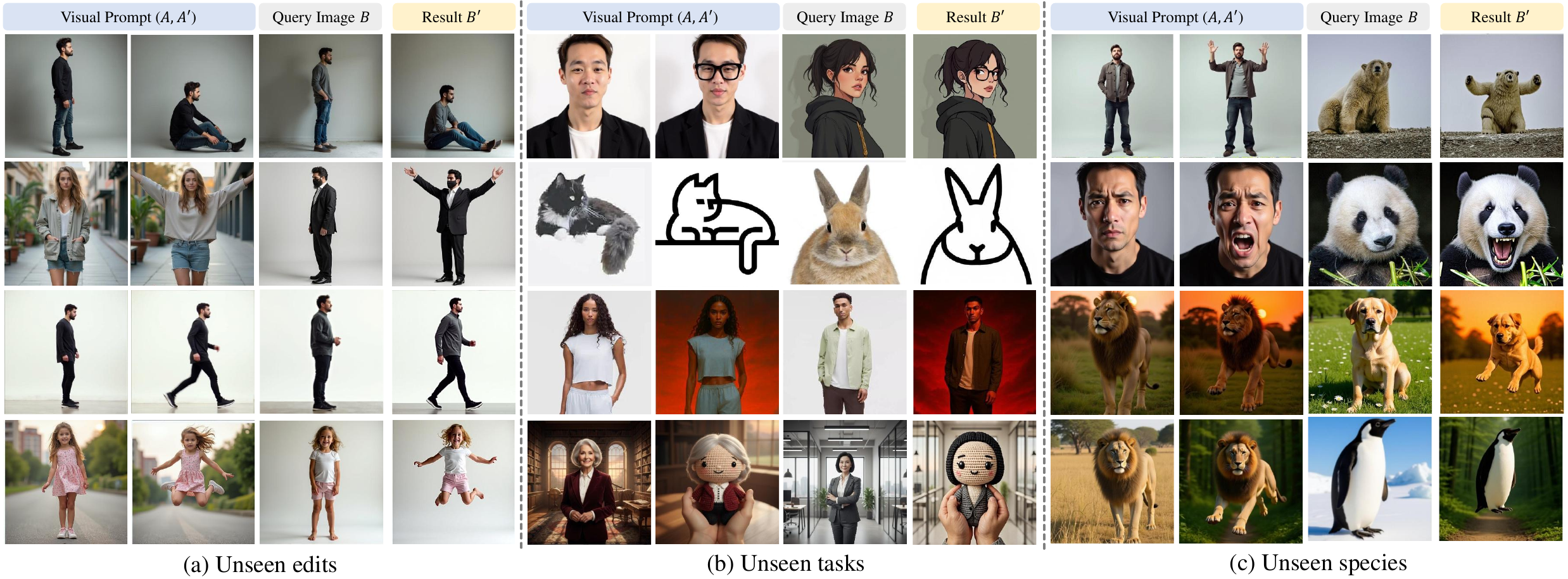}
    \vspace{-7 mm}
\caption{\textbf{Generalization of EditTransfer++.}
Qualitative results on (a) unseen edit variants, (b) unseen edit tasks, and (c) unseen species.
In all three settings, our method successfully follows the visual prompt for images beyond the training distribution, indicating strong generalization capability.}
    \label{fig:generalization}
    \vspace{-3 mm}
\end{figure*}

\subsection{Ablation Studies}
\label{subsec:AS}
We conduct an ablation over four proposed components.
Starting from the base model $M_b$ trained at a long-edge resolution of $1024$ without any of our strategies, we incrementally obtain
$M_1 = M_b + \text{TD}$ (text-decoupled training),
$M_2 = M_1 + \text{RF}$ (best--worst contrastive refinement),
$M_3 = M_2 + \text{DS}$ (condition compression),
and $M_4 = M_3 + \text{RU}$ (condition reuse), which is the full model.

\Paragraph{Impact of text-decoupled training.}
The text-decoupled strategy plays a critical role in enhancing the faithfulness with the visual prompt.
Building on the base model, we train a variant $M_1 = M_b + TD$ using null-text inputs and compare it with the base model $M_b$ trained with full textual prompts.
As shown in~\figref{ablation}(a), although $M_b$ produces reasonable outputs (\eg the side view edit in the first row) of the Relation252K~\cite{gong2025relationadapter} dataset, it fails to follow the visual prompt on unseen tasks in the second and third rows, exhibiting noticeably weaker alignment to the demonstrated transformations.  
Quantitative results in~\tabref{ablation} further confirm that applying text-decoupled training in $M_1$ leads to substantial gains in the CDS, especially on EditTransfer-Bench, demonstrating improved alignment with the visual prompt.

\Paragraph{Impact of best-worst contrastive refinement.}
We compare the outputs of $M_2$ and $M_1 = M_2 + \text{RF}$, both generated using the same seeds.  
As shown in~\figref{ablation}(b), incorporating RF effectively corrects previously suboptimal results, reducing visual artifacts and further enhancing faithfulness.   
For example, in the first row, the gesture produced without RF is incorrect, whereas incorporating RF rectifies the result and aligns it with the visual prompt.
Quantitative results in~\tabref{ablation} further support these observations, demonstrating improved generation consistency, as reflected by lower variance across samples.

\Paragraph{Impact of condition compression and reuse strategy.}
We first evaluate the efficiency–quality trade-off introduced by the proposed condition compression strategy.
Experiments are conducted under five resolution settings, with $(A, A')$ and $B$ downsampled to half ($512$) or a quarter ($256$) of the target output, which has a fixed long edge of $1024$ pixels.
As shown in~\figref{resolution} and~\tabref{downsample}, aggressively downsampling the visual prompts to $256$ greatly reduces inference time but leads to a noticeable loss in faithfulness, as reflected by the clear drop in CDS.
In contrast, setting $B$ to $256$ and $(A, A')$ to $512$ preserves visual alignment, as illustrated in the fifth column of~\figref{resolution}.
Therefore, we adopt this configuration and conduct experiments using $M_3 = M_2 + \text{DS}$.
As shown in~\tabref{ablation}, CDS of $M_3$ remains comparable, while its variance even exhibits a notable drop.
Building on $M_3$, the feature reuse strategy is applied to obtain $M_4 = M_3 + \text{RU}$, further reducing inference time.
Moreover, this strategy mitigates interference among condition features, yielding cleaner conditioning signals and improved performance, as shown in~\figref{ablation}(d) and~\tabref{ablation}.

%

\begin{figure}[!t]
    \includegraphics[width=\linewidth]{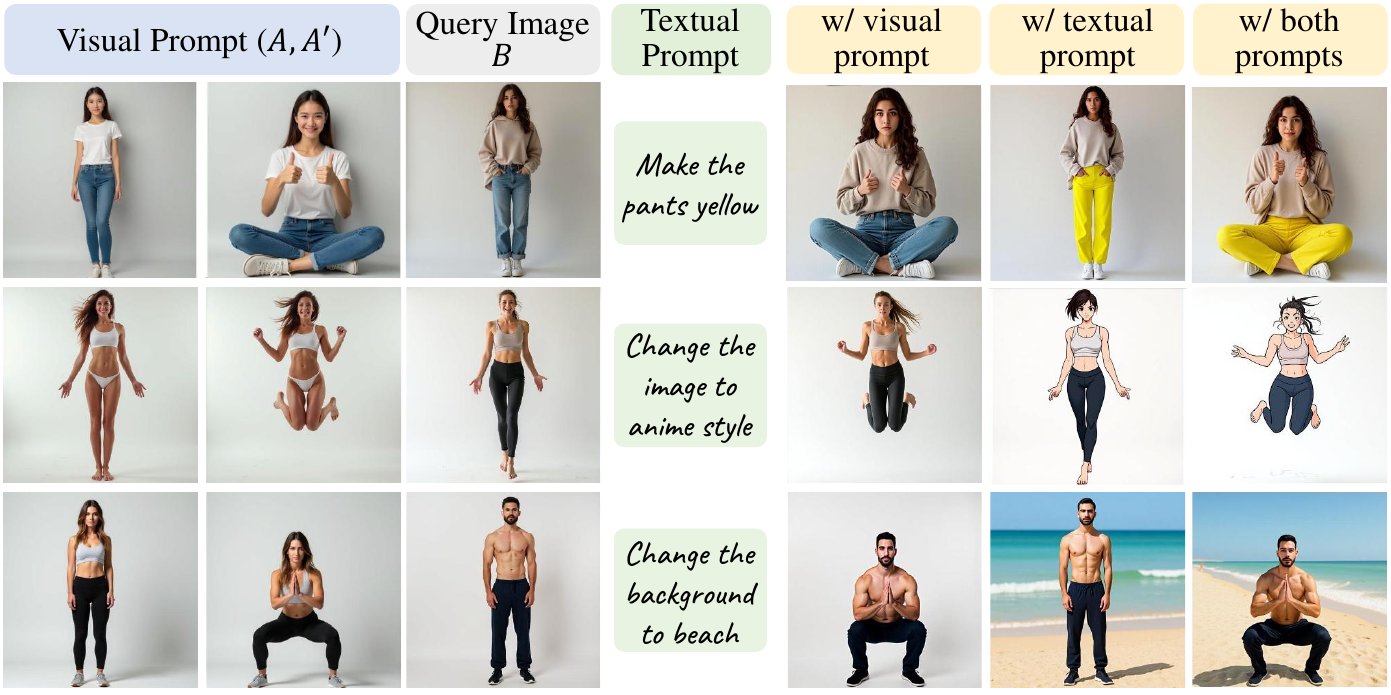}
    \vspace{-7 mm}
\caption{\textbf{Applications of EditTransfer++.}
Thanks to text-decoupled training, EditTransfer++ supports visual-prompt-only editing, text-only editing, and combined visual–textual control.
This enables flexible compositional edits by conditioning on visual prompts, textual instructions, or their combination.}
    \label{fig:composite}
    \vspace{-3 mm}
\end{figure}
\subsection{Discussions}
\label{subsec:discussion}

\Paragraph{Generalization capabilities.}
As illustrated in~\figref{generalization}, EditTransfer++ exhibits strong generalization ability across several scenarios that are not covered by the training distribution.
First, it generalizes to \emph{unseen variants of known edit types}, such as the ``sit down'' action, whose pose and contact patterns differ significantly from the training samples.
Second, it can handle \emph{novel edit types} that are absent from the training set, indicating that the model learns transferable editing behaviors rather than memorizing specific instances.
Third, it extends to \emph{cross-species edit transfer}, successfully applying transformations demonstrated on one species to visually distinct subjects from another, while preserving their identity and structure.

\Paragraph{Roles of text and visual prompts.}
The text-decoupled training strategy enables \emph{visual-prompt–only} editing, as shown in the fifth column of~\figref{composite}, where the model follows the demonstrated transformation without relying on textual instructions.
At the same time, the underlying T2I backbone retains its \emph{text-only} editing capability, allowing purely language-driven edits when no visual prompt is provided.
Moreover, EditTransfer++ naturally supports \emph{joint text–visual} conditioning, where textual and visual prompts are combined to specify the edit, providing flexible and fine-grained compositional control over the final result.

\section{Conclusions}
\label{sec:Conclusion}
We present \textbf{EditTransfer++}, a visual-prompt–guided image editing framework that couples a progressively structured training procedure with an efficient conditioning scheme. 
The proposed text-decoupled training stage encourages the backbone to learn transformations directly from visual evidence, while the best–worst contrastive refinement stabilizes the denoising trajectories and improves robustness to sampling seeds.
For better applicability and scalability, we introduce a condition compression and reuse strategy that reduces token redundancy, enabling high-resolution generation with lower memory usage and computational cost. 
To assess generalization in the edit transfer setting, we construct \textbf{EditTransfer-Bench}, which spans multiple editing types and includes both single- and multi-step edits. Across existing benchmarks and EditTransfer-Bench, EditTransfer++ consistently outperforms prior approaches, delivering more faithful visual-prompt adherence, more stable outputs, and faster inference.

Future work will extend EditTransfer++ to richer multi-condition editing scenarios beyond a single source–target pair and investigate its applicability to broader visual in-context learning tasks.

\bibliographystyle{IEEEtran}

\bibliography{main}

@String(PAMI = {IEEE Trans. Pattern Anal. Mach. Intell.})

@String(CVPR= {IEEE Conf. Comput. Vis. Pattern Recog.})

@String(ICCV= {Int. Conf. Comput. Vis.})

@String(NIPS= {Adv. Neural Inform. Process. Syst.})

@String(TOG= {ACM Trans. Graph.})

@String(TIP  = {IEEE Trans. Image Process.})

@String(ICLR = {Int. Conf. Learn. Represent.})

@String(AAAI = {AAAI})

@String(ICML = {ICML})

@String(SIGGRAPH={SIGGRAPH})

@String(PAMI  = {IEEE TPAMI})

@String(CVPR  = {CVPR})

@String(ICCV  = {ICCV})

@String(NIPS  = {NeurIPS})

@String(TOG   = {ACM TOG})

@String(TIP   = {IEEE TIP})

@String(ICLR  = {ICLR})

@String(ICICML = {ICICML})

@inproceedings{hertz2022prompt,
  title={Prompt-to-prompt image editing with cross attention control},
  author={Hertz, Amir and Mokady, Ron and Tenenbaum, Jay and Aberman, Kfir and Pritch, Yael and Cohen-Or, Daniel},
  booktitle=ICLR,
  year={2023},
}

@InProceedings{cao_2023_masactrl,
    author    = {Cao, Mingdeng and Wang, Xintao and Qi, Zhongang and Shan, Ying and Qie, Xiaohu and Zheng, Yinqiang},
    title     = {MasaCtrl: Tuning-Free Mutual Self-Attention Control for Consistent Image Synthesis and Editing},
    booktitle = ICCV,
    year      = {2023},
}

@inproceedings{brooks2023instructpix2pix,
  title={Instructpix2pix: Learning to follow image editing instructions},
  author={Brooks, Tim and Holynski, Aleksander and Efros, Alexei A},
  booktitle=CVPR,
  year={2023}
}

@inproceedings{zhang2025context,
  title={In-context edit: Enabling instructional image editing with in-context generation in large scale diffusion transformer},
  author={Zhang, Zechuan and Xie, Ji and Lu, Yu and Yang, Zongxin and Yang, Yi},
  booktitle=NIPS,
  year={2025}
}

@article{labs2025flux,
  title={FLUX. 1 Kontext: Flow Matching for In-Context Image Generation and Editing in Latent Space},
  author={Labs, Black Forest and Batifol, Stephen and Blattmann, Andreas and Boesel, Frederic and Consul, Saksham and Diagne, Cyril and Dockhorn, Tim and English, Jack and English, Zion and Esser, Patrick and others},
  journal={arXiv preprint arXiv:2506.15742},
  year={2025}
}

@inproceedings{wang2024taming,
  title={Taming rectified flow for inversion and editing},
  author={Wang, Jiangshan and Pu, Junfu and Qi, Zhongang and Guo, Jiayi and Ma, Yue and Huang, Nisha and Chen, Yuxin and Li, Xiu and Shan, Ying},
  booktitle=ICML,
  year={2023}
}

@inproceedings{feng2024dit4editdiffusiontransformerimage,
  title={Dit4edit: Diffusion transformer for image editing},
  author={Feng, Kunyu and Ma, Yue and Wang, Bingyuan and Qi, Chenyang and Chen, Haozhe and Chen, Qifeng and Wang, Zeyu},
  booktitle=AAAI,
  year={2025}
}

@inproceedings{avrahami2024stableflow,
    title={Stable Flow: Vital Layers for Training-Free Image Editing}, 
    author={Omri Avrahami and Or Patashnik and Ohad Fried and Egor Nemchinov and Kfir Aberman and Dani Lischinski and Daniel Cohen-Or},
    year={2025},
booktitle=CVPR
}

@inproceedings{zhang2024hive,
  title={Hive: Harnessing human feedback for instructional visual editing},
  author={Zhang, Shu and Yang, Xinyi and Feng, Yihao and Qin, Can and Chen, Chia-Chih and Yu, Ning and Chen, Zeyuan and Wang, Huan and Savarese, Silvio and Ermon, Stefano and others},
  booktitle=CVPR,
  year={2024}
}

@inproceedings{
zhang2023magicbrush,
title={MagicBrush: A Manually Annotated Dataset for Instruction-Guided Image Editing},
author={Kai Zhang and Lingbo Mo and Wenhu Chen and Huan Sun and Yu Su},
booktitle=NIPS,
year={2023}
}

@inproceedings{liu2024towards,
  title={Towards Understanding Cross and Self-Attention in Stable Diffusion for Text-Guided Image Editing},
  author={Liu, Bingyan and Wang, Chengyu and Cao, Tingfeng and Jia, Kui and Huang, Jun},
  booktitle=CVPR,
  year={2024}
}

@inproceedings{song2020denoising,
  title={Denoising diffusion implicit models},
  author={Song, Jiaming and Meng, Chenlin and Ermon, Stefano},
  year={2021},
  booktitle=ICLR
}

@inproceedings{ho2020denoising,
  title={Denoising diffusion probabilistic models},
  author={Ho, Jonathan and Jain, Ajay and Abbeel, Pieter},
  booktitle=NIPS,
  year={2020}
}

@inproceedings{rombach2022high,
  title={High-resolution image synthesis with latent diffusion models},
  author={Rombach, Robin and Blattmann, Andreas and Lorenz, Dominik and Esser, Patrick and Ommer, Bj{\"o}rn},
  booktitle=CVPR,
  year={2022}
}

@inproceedings{biswas2025PIXELS,
      title={PIXELS: Progressive Image Xemplar-based Editing with Latent Surgery}, 
      author={Shristi Das Biswas and Matthew Shreve and Xuelu Li and Prateek Singhal and Kaushik Roy},
      year={2025},
      booktitle=AAAI
}

@article{he2024freeedit,
  title={Freeedit: Mask-free reference-based image editing with multi-modal instruction},
  author={He, Runze and Ma, Kai and Huang, Linjiang and Huang, Shaofei and Gao, Jialin and Wei, Xiaoming and Dai, Jiao and Han, Jizhong and Liu, Si},
  journal=PAMI,
  year={2025}
}

@inproceedings{chen2024mimicbrush,
  title={Zero-shot image editing with reference imitation},
  author={Chen, Xi and Feng, Yutong and Chen, Mengting and Wang, Yiyang and Zhang, Shilong and Liu, Yu and Shen, Yujun and Zhao, Hengshuang},
    booktitle=NIPS,
  year={2025}
}

@INPROCEEDINGS{ChenSpecRef,
  author={Chen, Songyan and Huang, Jiancheng},
  booktitle=ICICML, 
  title={SpecRef: A Fast Training-free Baseline of Specific Reference-Condition Real Image Editing}, 
  year={2023}}

@INPROCEEDINGS{YangPaintbyExample,
  author={Yang, Binxin and Gu, Shuyang and Zhang, Bo and Zhang, Ting and Chen, Xuejin and Sun, Xiaoyan and Chen, Dong and Wen, Fang},
  booktitle=CVPR, 
  title={Paint by Example: Exemplar-based Image Editing with Diffusion Models}, 
  year={2023}}

@InProceedings{ChenAnyDoor,
    author    = {Chen, Xi and Huang, Lianghua and Liu, Yu and Shen, Yujun and Zhao, Deli and Zhao, Hengshuang},
    title     = {AnyDoor: Zero-shot Object-level Image Customization},
    booktitle = CVPR,
    year      = {2024},
}

@inproceedings{
  tan2024omini,
  title={OminiControl: Minimal and Universal Control for Diffusion Transformer},
  author={Zhenxiong Tan and Songhua Liu and Xingyi Yang and Qiaochu Xue and Xinchao Wang},
  year={2024},
booktitle=ICCV,
year={2025}
}

@article{lhhuang2024iclora,
  title={In-context lora for diffusion transformers},
  author={Huang, Lianghua and Wang, Wei and Wu, Zhi-Fan and Shi, Yupeng and Dou, Huanzhang and Liang, Chen and Feng, Yutong and Liu, Yu and Zhou, Jingren},
  journal={arXiv preprint arXiv:2410.23775},
  year={2024}
}

@INPROCEEDINGS{Wangspeak,
  author={Wang, Xinlong and Wang, Wen and Cao, Yue and Shen, Chunhua and Huang, Tiejun},
  booktitle=CVPR, 
  title={Images Speak in Images: A Generalist Painter for In-Context Visual Learning}, 
  year={2023}}

@inproceedings{Zhanggood,
 author = {Zhang, Yuanhan and Zhou, Kaiyang and Liu, Ziwei},
 booktitle = NIPS,
 title = {What Makes Good Examples for Visual In-Context Learning?},
 year = {2023}
}

@inproceedings{Barvisual,
author = {Bar, Amir and Gandelsman, Yossi and Darrell, Trevor and Globerson, Amir and Efros, Alexei A.},
title = {Visual prompting via image inpainting},
year = {2022},
booktitle =NIPS
}

@inproceedings{wang2023images,
  title={Images speak in images: A generalist painter for in-context visual learning},
  author={Wang, Xinlong and Wang, Wen and Cao, Yue and Shen, Chunhua and Huang, Tiejun},
  booktitle=CVPR,
  year={2023}
}

@inproceedings{gong2025relationadapter,
  title={RelationAdapter: Learning and Transferring Visual Relation with Diffusion Transformers},
  author={Gong, Yan and Song, Yiren and Li, Yicheng and Li, Chenglin and Zhang, Yin},
  booktitle=NIPS,
  year={2025}
}

@inproceedings{li2025visualcloze,
  title={VisualCloze: A Universal Image Generation Framework via Visual In-Context Learning},
  author={Li, Zhong-Yu and Du, Ruoyi and Yan, Juncheng and Zhuo, Le and Li, Zhen and Gao, Peng and Ma, Zhanyu and Cheng, Ming-Ming},
  booktitle=ICCV,
  year={2025}
}

@article{chen2025edit,
  title={Edit Transfer: Learning Image Editing via Vision In-Context Relations},
  author={Chen, Lan and Mao, Qi and Gu, Yuchao and Shou, Mike Zheng},
  journal={arXiv preprint arXiv:2503.13327},
  year={2025}
}

@inproceedings{yang2023imagebrush,
  title={Imagebrush: Learning visual in-context instructions for exemplar-based image manipulation},
  author={Yang, Yifan and Peng, Houwen and Shen, Yifei and Yang, Yuqing and Hu, Han and Qiu, Lili and Koike, Hideki and others},
  booktitle=NIPS,
  year={2023}
}

@misc{openai2024gpt4technicalreport,
      title={GPT-4o System Card}, 
      author={OpenAI},
      year={2024},
      eprint={2410.21276},
      archivePrefix={arXiv},
      primaryClass={cs.CL},
      url={https://arxiv.org/abs/2410.21276}, 
}

@inproceedings{zhou2025attentiondistillationunifiedapproach,
      title={Attention Distillation: A Unified Approach to Visual Characteristics Transfer}, 
      author={Yang Zhou and Xu Gao and Zichong Chen and Hui Huang},
      year={2025},
      booktitle=CVPR
}

@inproceedings{Alaluf2024transfer,
author = {Alaluf, Yuval and Garibi, Daniel and Patashnik, Or and Averbuch-Elor, Hadar and Cohen-Or, Daniel},
title = {Cross-Image Attention for Zero-Shot Appearance Transfer},
year = {2024},
booktitle = SIGGRAPH,
}

@INPROCEEDINGS{Zhucyclegan,
  author={Zhu, Jun-Yan and Park, Taesung and Isola, Phillip and Efros, Alexei A.},
  booktitle=ICCV, 
  title={Unpaired Image-to-Image Translation Using Cycle-Consistent Adversarial Networks}, 
  year={2017},}

@INPROCEEDINGS{Gatysstyle,
  author={Gatys, Leon A. and Ecker, Alexander S. and Bethge, Matthias},
  booktitle=CVPR, 
  title={Image Style Transfer Using Convolutional Neural Networks}, 
  year={2016}}

@inproceedings{peebles2023scalable,
  title={Scalable diffusion models with transformers},
  author={Peebles, William and Xie, Saining},
  booktitle=ICCV,
  year={2023}
}

@article{gal2022stylegan,
  title={Stylegan-nada: Clip-guided domain adaptation of image generators},
  author={Gal, Rinon and Patashnik, Or and Maron, Haggai and Bermano, Amit H and Chechik, Gal and Cohen-Or, Daniel},
  journal=TOG,
  year={2022},
}

@inproceedings{ruiz2023dreambooth,
  title={Dreambooth: Fine tuning text-to-image diffusion models for subject-driven generation},
  author={Ruiz, Nataniel and Li, Yuanzhen and Jampani, Varun and Pritch, Yael and Rubinstein, Michael and Aberman, Kfir},
  booktitle=CVPR,
  year={2023}
}

@inproceedings{zhang2025easycontrol,
  title={Easycontrol: Adding efficient and flexible control for diffusion transformer},
  author={Zhang, Yuxuan and Yuan, Yirui and Song, Yiren and Wang, Haofan and Liu, Jiaming},
  booktitle=ICCV,
  year={2025}
}

@inproceedings{liu2023unifying,
  title={Unifying image processing as visual prompting question answering},
  author={Liu, Yihao and Chen, Xiangyu and Ma, Xianzheng and Wang, Xintao and Zhou, Jiantao and Qiao, Yu and Dong, Chao},
  booktitle=ICML,
  year={2024}
}

@inproceedings{liu2022flow,
  title={Flow straight and fast: Learning to generate and transfer data with rectified flow},
  author={Liu, Xingchao and Gong, Chengyue and Liu, Qiang},
  booktitle=ICLR,
  year={2023}
}

@inproceedings{lipman2022flow,
  title={Flow matching for generative modeling},
  author={Lipman, Yaron and Chen, Ricky TQ and Ben-Hamu, Heli and Nickel, Maximilian and Le, Matt},
  booktitle=ICLR,
  year={2023}
}

@inproceedings{lu2025pairedit,
  title={PairEdit: Learning Semantic Variations for Exemplar-based Image Editing},
  author={Lu, Haoguang and Chen, Jiacheng and Yang, Zhenguo and Gnanha, Aurele Tohokantche and Wang, Fu Lee and Qing, Li and Mao, Xudong},
  booktitle=NIPS,
  year={2025}
}

@inproceedings{tumanyan2022splicing,
  title={Splicing vit features for semantic appearance transfer},
  author={Tumanyan, Narek and Bar-Tal, Omer and Bagon, Shai and Dekel, Tali},
  booktitle=CVPR,
  year={2022}
}

@misc{flux,
  title = {FLUX},
  author = {Black Forest Labs},
  year = {2024},
  url = {https://github.com/black-forest-labs/flux}
}

@inproceedings{wei2020multi,
  title={Multi-modality cross attention network for image and sentence matching},
  author={Wei, Xi and Zhang, Tianzhu and Li, Yan and Zhang, Yongdong and Wu, Feng},
  booktitle=CVPR,
  year={2020}
}

@inproceedings{loshchilov2017decoupled,
  title={Decoupled weight decay regularization},
  author={Loshchilov, Ilya and Hutter, Frank},
  booktitle=ICLR,
  year={2019}
}

@ARTICLE{11223644,
  author={Feng, Yunjian and Li, Jun and Zhou, MengChu},
  journal=TIP, 
  title={Instruction-Driven Multi-Weather Image Translation Based on a Large-Scale Image Editing Model}, 
  year={2025}}

@ARTICLE{10314461,
  author={Wu, Rongliang and Yu, Yingchen and Zhan, Fangneng and Zhang, Jiahui and Liao, Shengcai and Lu, Shijian},
  journal=TIP, 
  title={POCE: Pose-Controllable Expression Editing}, 
  year={2023}}

@ARTICLE{11218740,
  author={Xia, Tao and Zhang, Yudi and Liu, Ting and Zhang, Lei},
  journal=TIP, 
  title={Consistent Image Layout Editing With Diffusion Models}, 
  year={2025}}

@ARTICLE{10418856,
  author={Wang, Yaxiong and Wei, Yunchao and Qian, Xueming and Zhu, Li and Yang, Yi},
  journal=TIP, 
  title={ReGO: Reference-Guided Outpainting for Scenery Image}, 
  year={2024}}

@ARTICLE{11212802,
  author={Zhang, Yongle and Liu, Yimin and Fan, Hao and Hu, Ruotong and Zhang, Jian and Wu, Qiang},
  journal={IEEE Transactions on Image Processing}, 
  title={Consistent Image Inpainting With Pre-Perception and Cross-Perception Collaborative Processes}, 
  year={2025}}

@inproceedings{huang2025photodoodle,
  title={Photodoodle: Learning artistic image editing from few-shot pairwise data},
  author={Huang, Shijie and Song, Yiren and Zhang, Yuxuan and Guo, Hailong and Wang, Xueyin and Shou, Mike Zheng and Liu, Jiaming},
  booktitle=ICCV,
  year={2025}
}

@inproceedings{song2025omniconsistency,
  title={Omniconsistency: Learning style-agnostic consistency from paired stylization data},
  author={Song, Yiren and Liu, Cheng and Shou, Mike Zheng},
  booktitle=NIPS,
  year={2025}
}

@inproceedings{101145,
author = {Song, Yiren and Huang, Shijie and Yao, Chen and Ci, Hai and Ye, Xiaojun and Liu, Jiaming and Zhang, Yuxuan and Shou, Mike Zheng},
title = {ProcessPainter: Learning to draw from sequence data},
year = {2024},
booktitle = {SIGGRAPH Asia},
}

@inproceedings{zhang2025stable,
  title={Stable-hair: Real-world hair transfer via diffusion model},
  author={Zhang, Yuxuan and Zhang, Qing and Song, Yiren and Zhang, Jichao and Tang, Hao and Liu, Jiaming},
  booktitle=AAAI,
  year={2025}
}

@article{jiang2025personalized,
  title={Personalized Vision via Visual In-Context Learning},
  author={Jiang, Yuxin and Gu, Yuchao and Song, Yiren and Tsang, Ivor and Shou, Mike Zheng},
  journal={arXiv preprint arXiv:2509.25172},
  year={2025}
}

@article{song2025makeanything,
  title={Makeanything: Harnessing diffusion transformers for multi-domain procedural sequence generation},
  author={{ Y. Song, C. Liu, and  M. Z. Shou}},
  journal={arXiv preprint arXiv:2502.01572},
  year={2025}
}

\end{document}